\DeclareUnicodeCharacter{FB01}{fi}
\documentclass[10pt,journal,final,twocolumn,compsoc]{IEEEtran}
\pdfoutput=1
\usepackage{graphicx}
\usepackage{amsmath,amssymb} 
\usepackage{epsfig}
\usepackage{color}
\usepackage[ruled,vlined]{algorithm2e}
\usepackage{setspace}
\usepackage{bm}
\usepackage{lineno}
\usepackage{cases}
\usepackage{psfragx}
\usepackage{psfrag}
\usepackage{pifont}
\usepackage[table]{xcolor}
\usepackage{multirow}
\usepackage{makecell}
\usepackage{url}
\usepackage{stmaryrd}
\usepackage{arydshln}
\usepackage{mathrsfs}
\usepackage{verbatim}
\usepackage{enumerate}
\usepackage{lipsum}
\usepackage{booktabs}
\usepackage{threeparttable}
\usepackage[normalem]{ulem}
\usepackage{mathtools}

\newcommand{\eg}{e.g.}
\newcommand{\ie}{i.e.}

\newenvironment{shrinkeq}[1]
{ \bgroup
  \addtolength\abovedisplayshortskip{#1}
  \addtolength\abovedisplayskip{#1}
  \addtolength\belowdisplayshortskip{#1}
  \addtolength\belowdisplayskip{#1}}
{\egroup\ignorespacesafterend}

\newcommand{\HLblue}[1]{\textcolor{black}{#1}}

\newcommand{\HLredrevision}[1]{\textcolor{black}{#1}}

\SetCommentSty{mycommfont}
\definecolor{bleudefrance}{rgb}{0.19, 0.55, 0.91}
\definecolor{caribbeangreen}{rgb}{0.0, 0.8, 0.6}

\graphicspath{{./pdf/}{./tools/}{./pdf/supp/}}

\usepackage[top=0.5in, bottom=0.5in, left=0.5in, right=0.5in]{geometry}
\usepackage[pagebackref=false,breaklinks=true,colorlinks,linkcolor=black,anchorcolor=black,citecolor=black,urlcolor=black,bookmarks=false]{hyperref}

\ifCLASSOPTIONcompsoc
  \usepackage[nocompress]{cite}
\else
  \usepackage{cite}
\fi
\ifCLASSINFOpdf
\else
\fi
\ifCLASSOPTIONcompsoc
  \usepackage[caption=false,font=small,labelfont=sf,textfont=sf]{subfig}
\else
  \usepackage[caption=false,font=small]{subfig}
\fi
\begin{document}
%
\title{Recursive Least-Squares Estimator-Aided \\ Online Learning for Visual Tracking}
\author{Jin~Gao,
	Yan~Lu,
	Xiaojuan~Qi,
	Yutong~Kou,
	Bing~Li,
    Liang~Li,\\
    Shan~Yu,
	and Weiming~Hu,~\IEEEmembership{Senior~\HLblue{Member},~IEEE}
\IEEEcompsocitemizethanks{\IEEEcompsocthanksitem J. Gao, Y. Kou, B. Li, S. Yu and W. Hu are with the National Laboratory of Pattern Recognition, Institute of Automation, Chinese Academy of Sciences, Beijing 100190, P. R. China.\protect\\
	(E-mail: \{jin.gao, yutong.kou, bli, shan.yu, wmhu\}@nlpr.ia.ac.cn)
\IEEEcompsocthanksitem Y. Lu is with Microsoft Research Asia, Beijing 100080, P. R. China.\protect\\
	 (E-mail: yanlu@microsoft.com)
\IEEEcompsocthanksitem X. Qi is with the Department of Electrical and Electronic Engineering, The University of Hong Kong, Pokfulam Road, Hong Kong, P. R. China.\protect\\
	 (E-mail: xjqi@eee.hku.hk)
\IEEEcompsocthanksitem L. Li is with the Institute of Automation, Chinese Academy of Sciences, Beijing 100190, P. R. China. (E-mail: liang.li.brain@aliyun.com)
\IEEEcompsocthanksitem A preliminary version of this work appeared at CVPR 2020~\cite{Gao2020RLS-RTMDNet}.}
}

%
%

\markboth{Journal of \LaTeX\ Class Files,~Vol.~14, No.~8, August~2015}%
{Shell \MakeLowercase{\textit{et al.}}: Bare Demo of IEEEtran.cls for Computer Society Journals}
%



\IEEEtitleabstractindextext{%
\begin{abstract}
Tracking visual objects from a single initial exemplar in the testing phase has been broadly cast as a one-/few-shot problem, \ie, one-shot learning for initial adaptation and few-shot learning for online adaptation. The recent few-shot online adaptation methods incorporate the prior knowledge from large amounts of annotated training data via complex meta-learning optimization in the offline phase. This helps the online deep trackers to achieve fast adaptation and reduce overfitting risk in tracking. In this paper, we propose a simple yet effective recursive least-squares estimator-aided online learning approach for few-shot online adaptation without requiring offline training. It allows an in-built memory retention mechanism for the model to remember the knowledge about the object seen before, and thus the seen data can be safely removed from training. This also bears certain similarities to the emerging continual learning field in preventing catastrophic forgetting. This mechanism enables us to unveil the power of modern online deep trackers without incurring too much extra computational cost. We evaluate our approach based on two networks in the online learning families for tracking, \ie, multi-layer perceptrons in RT-MDNet and convolutional neural networks in DiMP. The consistent improvements on several challenging tracking benchmarks demonstrate its effectiveness and efficiency.
\end{abstract}

}

\maketitle

\IEEEdisplaynontitleabstractindextext

%
\IEEEpeerreviewmaketitle

\IEEEraisesectionheading{\section{Introduction}\label{sec:introduction}}

%
%
%
%
\IEEEPARstart{G}{iven} an arbitrary detected or annotated object of interest in the initial video frame, visual object tracking aims at {\it recognizing} and {\it localizing} other instances of the same object in subsequent frames to facilitate understanding how it moves through this video sequence. Recently this paradigm of tracking visual objects from a single initial exemplar in the testing phase has been broadly cast as a one-/few-shot meta-learning problem~\cite{Bertinetto2016SiameseFC, Bertinetto2016Learnet, Yang2017RFL, Park2018MetaSDNet, Li2018SiamRPN, Zhu2018DaSiamRPN, Yang2018MemTrack, Li2019SiamRPN++, Bhat2019DiMP,  Huang2019SSD-MAML, Choi2019MLT, Li2019GradNet, Zhang2019UpdateNet, Xu2020SiamFC++, Danelljan2020PrDiMP, Wang2020Retina-MAML, Jung2020MetaRTT, Dai2020LTMU, Huang2020GlobalTrack, Yang2020ROAM, Yang2021MemTrack}, which pushes forward the development of deep trackers and achieves unprecedented performance improvements.

It is well known that Deep Neural Networks (DNNs) with large capacity are typically data-hungry. However, a single initial exemplar in the initial frame can only provide limited training set for sequence-specific adaptation of a deep tracker. In contrast to some non-meta-learning-based pioneering arts which only fine-tune several light-weight classification network layers~\cite{Nam2016MDNet, Jung2018RTMDNet} or train them from scratch~\cite{Danelljan2019ATOM} to do the initial adaptation at the first frame for {\it recognizing}, many meta-learning-based deep trackers~\cite{Bertinetto2016SiameseFC, Bertinetto2016Learnet, Park2018MetaSDNet, Li2018SiamRPN, Zhu2018DaSiamRPN, Li2019SiamRPN++, Bhat2019DiMP, Huang2019SSD-MAML, Xu2020SiamFC++, Danelljan2020PrDiMP, Wang2020Retina-MAML, Jung2020MetaRTT, Huang2020GlobalTrack} are proposed to speed up this initial adaptation process and reduce the risk of overfitting. The major insight is to inject additional prior knowledge into the meta-learning-based one-shot adaptation process to achieve explicit model initialization $\bm{\theta}_{\textrm{init}}^0$ through solving massive amounts of small meta-learning tasks based on the single initial exemplar in the offline learning process.

It is also noteworthy that, in the visual tracking testing phase, it is possible to augment the initial limited training set to update the trackers online by exploiting the position predictions of the object in subsequent frames as additional samples. Despite the fact that introducing imperfect predictions into the learning process may have the effect of accumulating errors and leading to tracking drift, the conditions under which a visual tracking system setup is performed are somehow less extreme~\cite{Bertinetto2019} in comparison with one-/few-shot classification. As the number of arriving samples increases, one may find an ideal and straightforward way to adapt the tracking models to new environments online and avoid overfitting, which is fine-tuning the model parameters over the whole increasingly larger sample set at each update step. However, in the circumstances of online deep trackers, it may encounter a significant computational overhead due to the requirement of more gradient descent iterations to achieve good convergence, which prohibits high tracking speed. In practice, a handful of samples are maintained during the whole testing phase in a way similar to the {\it sliding-window} strategy with fixed-window size. Again, to speed up this online adaptation and reduce the risk of overfitting, some of the aforementioned meta-learning-based one-shot initial adaptation methods are extended~\cite{Huang2019SSD-MAML, Wang2020Retina-MAML, Jung2020MetaRTT}, and some novel meta-learners are also carefully designed to offline-train updaters~\cite{Yang2017RFL, Yang2018MemTrack, Choi2019MLT, Li2019GradNet, Zhang2019UpdateNet, Yang2020ROAM, Yang2021MemTrack} for effective online adaptation, namely few-shot online adaptation. 

Despite the success of both the meta-learning-based one-shot initial adaptation and few-shot online adaptation strategies in visual tracking to avoid overfitting and reduce memory or computation requirements based on fast domain adaptation, complex optimization is required in the offline training phase -- especially for the latter adaptation strategy.  Specifically, both of them incorporate the prior knowledge from large amounts of annotated training data for fast domain adaptation, \eg, the training splits of \textit{ImageNet DET} and \textit{VID}~\cite{Russakovsky2015ImageNet}, \textit{MS-COCO}~\cite{Lin2014COCO}, \textit{GOT10k}~\cite{Huang2019GOT}, \textit{TrackingNet}~\cite{Muller2018TrackingNet} and \textit{LaSOT}~\cite{Fan2019LaSOT}. In this paper, we take a different perspective than the above methods for addressing the few-shot online adaptation problem, and propose a simple yet effective online adaptation method based on recursive least-squares (RLS) estimation~\cite{Haykin2002AFT}. The motivation behind this is that we find it is feasible to sidestep the computational overhead of fine-tuning over the whole gradually enlarged sample set by exploring the RLS's ability to ``not forget" the old knowledge about the object after discarding the old data. Our approach can be treated as a widely applicable and integrable module for existing deep trackers without incurring too much extra computational cost in the sense that it improves the plain gradient descent optimization over the {\it sliding-window} data for tracking whereas neither the time to perform optimizations nor the storage consumption significantly increase.

\begin{figure}[!t]
\captionsetup{font={small}}
\centering
\subfloat{\includegraphics[width=0.99\linewidth]{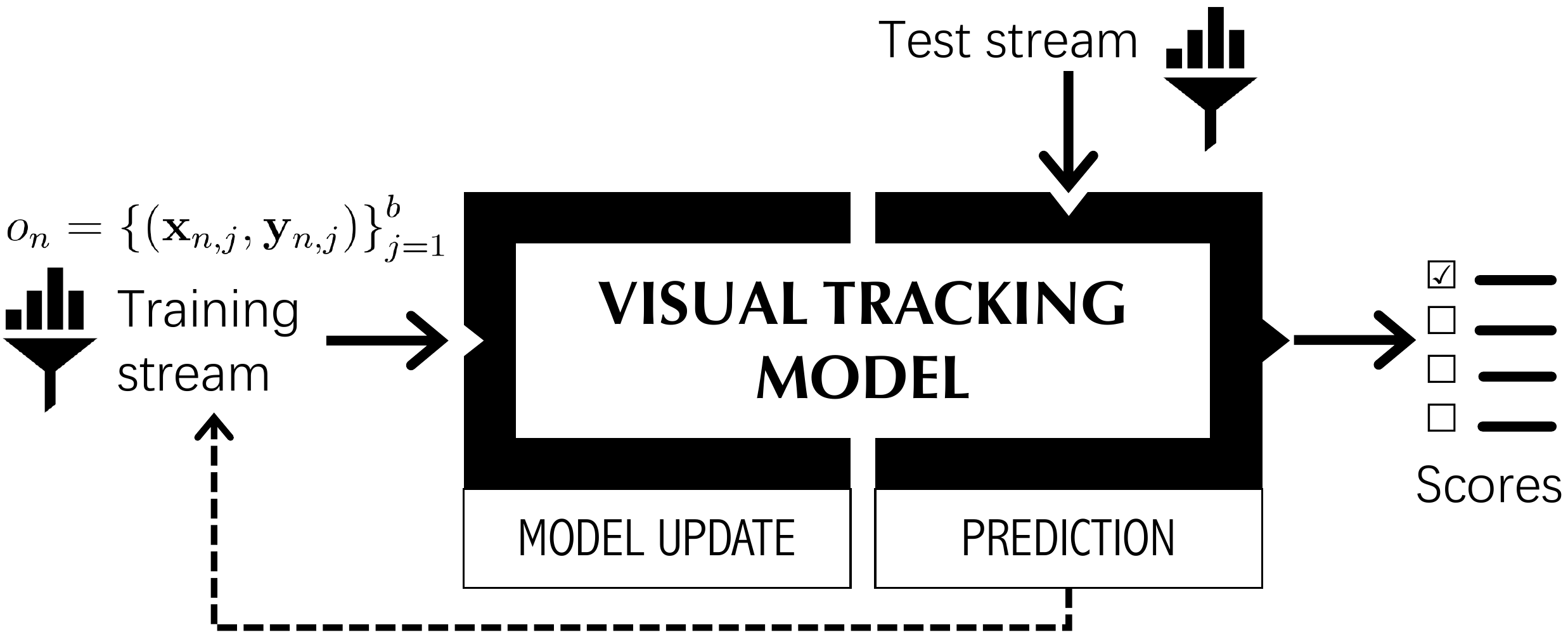}
\label{OnlineTracking}}
\caption{Online learning in visual tracking. Here we denote the collected $b$ training samples at time index $n$ as $o_n = \left\{\left(\mathbf{x}_{n,j},  \mathbf{y}_{n,j}\right )\right\}_{j=1}^b$.} \label{fig:OnlineTracking}
\end{figure}

In other words, we propose to use RLS estimation to aid the online adaptation. Our aim is to retain past memory while updating tracking models sequentially based on a handful of samples maintained at each update step. This resembles the emerging continual learning literature~\cite{Kirkpatrick2017EWC, He2018CAB, Zeng2019OWM, Parisi2018CLReview}, which has recently proved valuable in numerous typical supervised learning and reinforcement learning-based tasks where some static models are offline trained based on continual learning with fixed volumes of data. The few-shot online adaptation is also closely related to generic online learning, which basically aims at updating the predictor for future data at each time index while a continuous stream of data become available in a sequential order. It sometimes need dynamically adapting to new patterns in a stream when the data itself change from time to time, which is just right for visual tracking as shown in Figure~\ref{fig:OnlineTracking}. So our approach departs from the prior arts in that we are the first to improve on the online learning procedure by incorporating memory retention in the spirit of continual learning in humans without catastrophic forgetting.

The concrete implementation to demonstrate the effectiveness of our approach is based on the fact that the calculations in deep networks, no matter multi-layer perceptrons (MLPs) or convolutional neural networks (ConvNets), can be quickly performed using fast linear algebra routines by organizing the network parameters in matrices. We study two networks in the online learning families for tracking: the MLPs-based as in RT-MDNet~\cite{Jung2018RTMDNet}, and the ConvNets-based as in DiMP~\cite{Bhat2019DiMP}. The reason for choosing these baselines is that they both use the {\it sliding-window} strategy for online learning classification models and {\it recognizing} objects in tracking, and the exploited optimization methods only enable the learning to converge to a {\it local} point with respect to the fixed-sized sample set at each update step. Instead, our improved optimization method in a recursive fashion can enable the learning to approximately converge to a {\it global} point with respect to all the historical training samples ever seen including the discarded old data. What's more, we do not rely on offline training to improve the tracking performance, and all the validations are based on the off-the-shelf models of the baselines in their papers.

In summary, through network online learning we investigate a continual-learning-inspired online learning approach with RLS estimator-aided online adaptation in tracking, which is orthogonal to the investigation of meta-learning approaches. We believe the observations in this paper will improve the understandings of network online learning for visual tracking, and we expect our simple-to-implement approach will further advance the few-shot online adaptation research. Our improved code and raw results are released at \url{https://github.com/Amgao/RLS-OnlineTrack}.

%
%

\section{Related work}\label{sec:relatedwork}

\subsection{Online Learning in Visual Tracking}

Online learning has been an important part of visual tracking for about ten years since the classic IVT tracker~\cite{Ross2008IVT} was first proposed in 2008. The subsequent research has been concentrating on online robust classifier construction, including MIL~\cite{Babenko2011MIL}, Struck~\cite{Hare2016Struck}, MEEM~\cite{Zhang2014MEEM}, TGPR~\cite{Gao2020TGPR}, DCF-based approaches~\cite{Henriques2015KCF, Danelljan2017DSST, Danelljan2015SRDCF, Galoogahi2017BACF, YLi2019ARCF, Danelljan2016CCOT, Danelljan2017ECO, Xu2019LADCF, Li2018STRCF, Xu2019GFS-DCF, YLi2020AutoTrack}, deep classifiers~\cite{Nam2016MDNet, Jung2018RTMDNet, Danelljan2019ATOM}, and some recent long-term trackers~\cite{Yang2014Reacquisition, Dai2020LTMU}, just to name a few. This is very different from the typical supervised learning tasks training with fixed volumes of data, \eg, image classification~\cite{Simonyan2015VGG} and object detection~\cite{He2017MaskRCNN}.

In general, adapting to new data continually may give the online tracking models a tremendous advantage over the Siamese-style trackers~\cite{Bertinetto2016SiameseFC, Bertinetto2016Learnet, Li2018SiamRPN, Zhu2018DaSiamRPN, Li2019SiamRPN++, Xu2020SiamFC++, Chen2020SiamBAN} with online learning abandoned in discriminating background distractors. This instance-level discrimination power largely relies on the convergence rate of optimization algorithms, especially within limited optimization iterations allowed in the update stage for the sake of tracking speed. The challenges may stem from the learning rate selection, model initialization, non-convex cost function, and so on~\cite{Ruder2017GDReview}. 
Besides providing a good model initialization to account for the convergence challenges in the update stage as the aforementioned meta-learning-based one-shot initial adaptation methods~\cite{Park2018MetaSDNet, Bhat2019DiMP, Huang2019SSD-MAML, Danelljan2020PrDiMP, Wang2020Retina-MAML, Jung2020MetaRTT} have done, some novel LSTM model-based~\cite{Yang2017RFL, Yang2018MemTrack, Yang2020ROAM, Yang2021MemTrack}, ConvNet model-based~\cite{Zhang2019UpdateNet}, and gradient-based~\cite{Choi2019MLT, Li2019GradNet} meta-learners are proposed to efficiently adapt tracking models to appearance variations. Moreover, the Conjugate Gradient~\cite{Shewchuk1994CG} methodology-based online optimizers~\cite{Danelljan2016CCOT, Danelljan2017ECO, Danelljan2019ATOM} or Steepest Descent methodology-based ones~\cite{Bhat2019DiMP} are also carefully designed to improve convergence while online learning.

In addition, despite the benefits of improved instance-level discrimination power with frequent update, the excessive update with limited memory size due to the discarding of old data may result in overfitting to recent training samples, deteriorating the robustness. Being robust is crucial for increasing the flexibility in tackling the different distortions of the object appearance over time (\eg, intrinsic object scale and pose variations, and variations caused by extrinsic illumination change, camera motion and occlusion). In other words, the plain online learning in visual tracking, \eg, the standard gradient descent (GD) or its momentum-based stochastic twin (SGD)~\cite{Qian1999Momentum} in the earlier online trackers~\cite{Danelljan2015SRDCF, Nam2016MDNet, Jung2018RTMDNet}, faces challenges in meeting both the instance-level discrimination and robustness demands at the same time, as they are sometimes contradictory~\cite{Wang2019SPM}. Apart from the aforementioned meta-learning-based few-shot online adaptation methods, some online trackers widely apply the moderately infrequent and underfitted update~\cite{Danelljan2016CCOT, Danelljan2017ECO, Danelljan2019ATOM, Bhat2019DiMP, Danelljan2020PrDiMP}, passive-aggressive update~\cite{Li2018STRCF, Xu2019LADCF, Xu2019GFS-DCF, YLi2020AutoTrack}, or long short-term complementary update~\cite{Nam2016MDNet, Jung2018RTMDNet} settings in the update stage to relieve suffering from overfitting to recent training samples, or even corrupted samples.

These observations inspire the design of our proposed RLS estimator-aided online learning approach for visual tracking, which not only has instance-level discrimination power in case of background distractors as other online trackers does, but also can enhance the robustness ability thanks to the memory retention despite the discarding of old data. 

\subsection{One-/Few-shot Learning in Visual Tracking}
It is well known that humans can effortlessly learn novel concepts after only one or a few exposures by exploiting the knowledge primitives accrued in a lifetime. Inspired by this one-/few-shot learning ability of humans, there has been a recent resurgence of interest in two kinds of machine learning problems, \ie, (a) learning a meta-level knowledge across a large number of distinct training tasks to rapidly generalise to new concepts with small training data, which is termed as meta-learning or learning-to-learn (\eg,~\cite{Koch2015MetaSiamese, Bertinetto2016Learnet, Finn2017MAML}), and (b) learning consecutive tasks to continually adapt to new data without forgetting how to perform previously trained tasks, namely continual learning (\eg,~\cite{Kirkpatrick2017EWC, He2018CAB, Zeng2019OWM, Parisi2018CLReview}). Since the limited-data regime that characterises the visual tracking setup has attracted much attention recently, a promising research direction is the one set by applying one-/few-shot learning paradigm to visual tracking.

Siamese networks are firstly exploited to formulate the one-shot object recognition task as image matching via learning feature representations that preserve the class neighborhood structure and then inducing the $l_1$ component-wise distance metric to join the two Siamese twins for similarity computation~\cite{Koch2015MetaSiamese}. This metric-learning-based meta-learning approach is further extended to be capable of predicting network parameters in~\cite{Bertinetto2016Learnet} and two more distance metrics are evaluated, \ie, the inner-product and the Euclidean distance. In~\cite{Bertinetto2016SiameseFC}, a Siamese architecture that is fully-convolutional with respect to the search image by obtaining the convolution parameters from the template image, namely SiamFC, is designed for dense and efficient matching-based visual tracking in a one-shot manner. This efficiency essentially benefits from encoding the inner-product similarity between the template image and each spatial location in the search image. Following the SiamFC paradigm, many Siamese network-based one-shot learning methods for initial adaptation~\cite{Li2018SiamRPN, Zhu2018DaSiamRPN, Li2019SiamRPN++, Bhat2019DiMP, Xu2020SiamFC++, Huang2020GlobalTrack, Danelljan2020PrDiMP} and few-shot learning methods for online adaptation~\cite{Yang2017RFL, Yang2018MemTrack, Choi2019MLT, Li2019GradNet, Zhang2019UpdateNet, Yang2021MemTrack} in visual tracking are studied.

MAML~\cite{Finn2017MAML, antoniou2019how}, a model-agnostic meta-learning approach, aims to help the network to learn a set of good initialization parameters that are suitable for fine-tuning to produce good generalization performance on a new task efficiently. Since it is compatible with different models trained with gradient descent without changing their architectures, some recent works~\cite{Park2018MetaSDNet, Huang2019SSD-MAML, Wang2020Retina-MAML, Jung2020MetaRTT} formulate tracking in a MAML-based one-shot initial adaptation or few-shot online adaptation framework. They either improve the existing trackers~\cite{Park2018MetaSDNet, Jung2020MetaRTT} or directly convert a modern object detector into a tracker~\cite{Huang2019SSD-MAML, Wang2020Retina-MAML}, leading to improvements in tracking speed and robustness, or resusability of the advancement in object detection.

Continual learning, an attractive route to artificial general intelligence, has been demonstrated to be capable of solving the classic image classification problem~\cite{Kirkpatrick2017EWC, He2018CAB, Zeng2019OWM} or Atari 2600 games~\cite{Kirkpatrick2017EWC} through sequential learning. Besides overcoming catastrophic forgetting in the offline training phase like these earlier works, a preliminary version of this paper published in CVPR’20~\cite{Gao2020RLS-RTMDNet} is the first to improve on the online learning procedure by incorporating memory retention and demonstrate its success in visual tracking. It has theoretically and experimentally demonstrated retaining past memory while updating tracking models sequentially based on RLS estimation can tackle catastrophic forgetting incurred by plain SGD in the online learning of MLP layers in RT-MDNet~\cite{Jung2018RTMDNet}. The present work goes further to investigate RLS's ability of a convolutional layer for online learning improvement in tracking. To this end, a recent state-of-the-art tracker DiMP~\cite{Bhat2019DiMP} is exploited as another baseline,  which enables us to demonstrate the effectiveness of our approach on the more challenging \textit{VOT2018/2019} benchmarks~\cite{Kristan2018VOTconf, Kristan2019VOTconf}, the large-scale short-term \textit{TrackingNet-test}~\cite{Muller2018TrackingNet} and \textit{GOT10k-test}~\cite{Huang2019GOT} tracking benchmarks, and the long-term \textit{LaSOT-test}~\cite{Fan2019LaSOT}, \textit{OxUvA-dev}~\cite{Valmadre2018OxUvA} and \textit{TLP}~\cite{Moudgil2019TLP} benchmarks. The consistent improvements on several challenging benchmarks against the plain GD prove our online adaptation method's efficiency without additional offline training and too much tedious work on parameter adjusting. In the supplementary material, we also provide additional technical details to explain, for instance, how to extend the derivation of normal equations for MLP in~\cite{Gao2020RLS-RTMDNet} to handle non-linear activation functions and how the derivation of normal equations for MLP in~\cite{Gao2020RLS-RTMDNet} can be easily transplanted to the case of cross-entropy loss.

\section{RLS Estimator-Aided Online Learning with Memory Retention}
\label{Sec:RLS}

Nowadays, online learning in visual tracking is increasingly related to online learning network parameters from a carefully designed and time-varying sampling set~\cite{Jung2018RTMDNet, Bhat2019DiMP}. By organizing the network parameters in matrices, the calculations in the networks can be quickly performed using fast linear algebra routines. In this section, a simple least-squares estimation (LSE) problem is firstly investigated in matrix form to facilitate the interpretation of our approach.

\subsection{LSE in Online Learning}
\label{Sec:Problem}
Consider the time-varying training set $\left\{\bm{X}(n), \bm{Y}(n)\right\}$ consisting of the observable input samples and desired responses within all of the time span until the time index $n$, the method of least squares can be used to estimate a set of weighting coefficients in $\mathbf{W}_n$ at the time index $n$ in the online learning procedure, such that the following {\it cost function} is minimized:
\begin{shrinkeq}{-1ex}
\begin{align}
\label{equ:costfunctionLS} \mathscr{L}(\mathbf{W}_n) = \left\Vert \bm{\varLambda}(n)\left(\bm{Y}(n) - \bm{X}(n)\mathbf{W}_n^{\top}\right)\right\Vert^2\! + \delta\beta^n\left\Vert \mathbf{W}_n\right\Vert^2\!,
\end{align}
\end{shrinkeq}
where the left squared error term is defined based on the $l_2$ norm of the {\it residual} vectors, and the regularizing term, controlled by a positive constant $\delta$, is included to stabilize the solution to the recursive formulation of this LSE problem~\cite{Haykin2002AFT, Moustakides1997} as shown in Section~\ref{Sec:RLS-Formulation}.
Here we denote $\bm{X}(n)$, $\bm{Y}(n)$ and $\bm{\varLambda}(n)$ in a {\it block} manner. The sampling set $\left\{\bm{X}(n), \bm{Y}(n)\right\}$ has growing dimensions in the number of data {\it blocks} in rows (growing-window),
\begin{shrinkeq}{-1ex}
\begin{align}
\label{equ:XY} \bm{X}(n) = \begin{pmatrix}
\bm{X}_1 \\
\vdots \\
\bm{X}_n\\
\end{pmatrix} \in \mathbb{R}^{nb \times p}~, ~~~ \bm{Y}(n) = \begin{pmatrix}
\bm{Y}_1 \\
\vdots \\
\bm{Y}_n\\
\end{pmatrix} \in \mathbb{R}^{nb \times q}~,
\end{align}
\end{shrinkeq}
with $\bm{X}_i \!\in\! \mathbb{R}^{b \times p}, \bm{Y}_i \!\in\! \mathbb{R}^{b \times q}$ and $i = 1,2,\cdots,n$. Here $b$, $p$ and $q$ are to denote the {\it block} size and the dimensions of the input and output for each sample. The diagonal weighting matrix 
\begin{shrinkeq}{-1ex}
\begin{align}
\label{equ:Lambda} \bm{\varLambda}(n) = diag\left(\sqrt{\beta}^{n-1}\mathbf{I}_b,\cdots, \sqrt{\beta}\mathbf{I}_b, \mathbf{I}_b\right)
\end{align}
\end{shrinkeq}
is to diminish the importance of those previous observations (rows) with $0 < \beta \leq 1$, which can measure the memory of the algorithm. The regularizing term is also reformulated to make its beneficial effect forgotten with time for $\beta$ less than unity.

According to the method of LSE~\cite{Haykin2002AFT}, the optimum value of $\widehat{\mathbf{W}}_n$ at the time index $n$, for which Eq.~\eqref{equ:costfunctionLS} attains its minimum value, is defined by the normal equations. Before that, we can rewrite the {\it cost function} in Eq.~\eqref{equ:costfunctionLS} as
\begin{shrinkeq}{-1ex}
\begin{gather}
\label{equ:costfunctionLSnew2} \mathscr{L}(\mathbf{W}_n) \!=\! \mathop{\sum}\limits _{i=1}^{n}\beta^{n-i}\!\left(\frac{1}{b}\left\Vert \bm{Y}_i - \bm{X}_i\mathbf{W}_n^{\top}\right\Vert^2\right) \!+\! \frac{\delta}{b}\beta^n\left\Vert \mathbf{W}_n\right\Vert^2~\!\!.
\end{gather}
\end{shrinkeq}
Solving for $\mathbf{W}_n$ for which $\nabla _{\mathbf{W}_n}\mathscr{L}$ is zero, we may write the normal equations as follows in matrix form:
\begin{shrinkeq}{-1ex}
\begin{align}
\label{equ:normalequ_memo1} \widehat{\mathbf{W}}_n = \mathbf{Z}_n\bm{\Phi}_n^{-1}~,
\end{align}
\end{shrinkeq}
where $\mathbf{Z}_n$ and $\bm{\Phi}_n$ are defined respectively by Eqs.~\eqref{equ:normalequ_memo2} and \eqref{equ:normalequ_memo3}, \ie, the time-average cross-correlation matrix $\mathbf{Z}_n$ between the desired output and the input is shown by the formula
\begin{shrinkeq}{-1ex}
\begin{align}
\label{equ:normalequ_memo2} \mathbf{Z}_n = \mathop{\sum}\limits _{i=1}^{n}\beta^{n-i}\bm{Y}_i^{\top}\bm{X}_i~,
\end{align}
\end{shrinkeq}
and the time-average cross-correlation matrix $\bm{\Phi}_n$ of the input including the regularizing term is defined by
\begin{shrinkeq}{-1ex}
\begin{align}
\label{equ:normalequ_memo3} \bm{\Phi}_n = \mathop{\sum}\limits _{i=1}^{n}\beta^{n-i}\bm{X}_i^{\top}\bm{X}_i + \delta\beta^{n}\mathbf{I}~.
\end{align}
\end{shrinkeq}

In practice, SGD or its mini-batch version (MBSGD) is commonly used to optimize the parameters in $\mathbf{W}_n$ to obtain an approximate optimum value $\widehat{\mathbf{W}}_{n}^{appr}$ for problems similar to Eq.~\eqref{equ:costfunctionLS} or \eqref{equ:costfunctionLSnew2}. The initialization of this optimization at the time index $n$ is commonly set to $\widehat{\mathbf{W}}_{n-1}^{appr}$, which is the approximate optimum value at the time index $n-1$. This iterative optimization procedure for online learning can not only avoid the computationally expensive inverse operation in Eq.~\eqref{equ:normalequ_memo1} for high-dimensional input data, but also facilitate the training of networks in deep learning by allowing processing the continuously and increasingly growing large dataset with limited GPU memory size. As the number of arriving {\it blocks} increases, however, fine-tuning the parameters over the whole increasingly larger training set $\left\{\bm{X}(n), \bm{Y}(n)\right\}$ will give rise to more gradient descent iterations to achieve good convergence, which leads to a significant computational overhead and thus prohibits high tracking speed. Moreover, the growing training set results in growing storage consumption. 

In order to make a compromise between the optimization and tracking speed, traditional online learning in visual tracking commonly uses a carefully designed {\it sliding-window} strategy with fixed-window size to incorporate the new data segments and also remove the influence of the obsolete data. Similar to this traditional online learning paradigm, a strategy to solve Eq.~\eqref{equ:costfunctionLS} can be derived as follows. First, if we denote $s$ as the number of {\it blocks} of the fixed-window size, then for $n \geq s$, the fixed-window data can be written from the growing-window data given in Eq.~\eqref{equ:XY} by discarding the oldest $n-s$ {\it blocks} of data, \ie,
\begin{shrinkeq}{-1ex}
\begin{align}
\label{equ:XYnew} \bm{X}(n) = \begin{pmatrix}
\bm{X}_{n-s+1} \\
\vdots \\
\bm{X}_n\\
\end{pmatrix} ~, ~~~ \bm{Y}(n) = \begin{pmatrix}
\bm{Y}_{n-s+1} \\
\vdots \\
\bm{Y}_n\\
\end{pmatrix}~.
\end{align}
\end{shrinkeq}
Second, the approximate optimum value $\widehat{\mathbf{W}}_n^{appr}$ can be obtained by minimizing Eq.~\eqref{equ:costfunctionLS} based on some optimization strategies, \eg, standard gradient descent or its stochastic twin SGD as in RT-MDNet~\cite{Jung2018RTMDNet}, and Steepest Descent (SD) methodology as in DiMP~\cite{Bhat2019DiMP}. The initialization at the time index $n$ is also set to $\widehat{\mathbf{W}}_{n-1}^{appr}$. For instance, the standard batch gradient descent (BGD) can be used to optimize $\mathbf{W}_n$ based on the following iterations:
\begin{shrinkeq}{-1ex}
\begin{align}
\label{equ:derivativeG} \mathbf{W}_n^{iter+1} &= \mathbf{W}_n^{iter} - \eta \left(\mathbf{W}_n^{iter}\bm{\Phi}_n - \mathbf{Z}_n\right)~,\\
\label{equ:updateG} \mathbf{W}_n^{0} &= \widehat{\mathbf{W}}_{n-1}^{appr}~.
\end{align}
\end{shrinkeq}
Here $\eta$ is a pre-defined learning rate, and $\bm{\Phi}_n$ and $\mathbf{Z}_n$ are obtained by summing over the fixed-window data.

Despite the success of this {\it sliding-window} strategy with fixed-window size, a limitation of historical memory loss prevents unveiling the power of online deep tracking in~\cite{Jung2018RTMDNet, Bhat2019DiMP}, because this strategy will potentially be faced with the problem of being prone to overfitting to the recent fixed-window data especially when the size of the window is limited. This raises an interesting question that we discuss in the next subsection: Can we have a more elegant path to retain the memory while neither increasing the time to perform optimizations nor the storage consumption significantly? We hope our work and the presented online learning approach allow researchers to focus on the still unsolved critical challenges of online deep tracking.

\subsection{Recursive Formulation with Memory Retention}
\label{Sec:RLS-Formulation}
This section investigates the relationship between $\widehat{\mathbf{W}}_{n}$ and $\widehat{\mathbf{W}}_{n-1}$ at the consecutive time indices and presents an interesting alternative to solving Eq.~\eqref{equ:costfunctionLS} in a recursive fashion~\cite{Haykin2002AFT}. In contrast to the aggressive online learning in the previous {\it sliding-window} strategy, the characteristic of this recursive online learning can spontaneously reduce the risk of overfitting thanks to the memory retention though the oldest {\it blocks} of data are discarded. In other words, the catastrophic forgetting of old knowledge about discriminating the discarded training samples during the independent optimization over the newly updated fixed-window sample set in Eqs.~\eqref{equ:derivativeG} and \eqref{equ:updateG} is mitigated. 

To simplify the derivation, we only consider one sample pair in each {\it block}, which means $b=1$ in Eq.~\eqref{equ:costfunctionLS}. Let $\bm{X}_i = \mathbf{x}_i^{\top}$ and $\bm{Y}_i = \mathbf{y}_i^{\top}$, the {\it cost function} of Eq.~\eqref{equ:costfunctionLS} will degrade to 
\begin{shrinkeq}{-1ex}
\begin{align}
\label{equ:costfunctionLSnew} \mathscr{L}(\mathbf{W}_n) = \mathop{\sum}\limits _{i=1}^{n}\beta^{n-i}\left\Vert \mathbf{y}_i - \mathbf{W}_n\mathbf{x}_i\right\Vert^2 + \delta\beta^n\left\Vert \mathbf{W}_n\right\Vert^2~,
\end{align}
\end{shrinkeq}
and the cross-correlation matrices in Eq.~\eqref{equ:normalequ_memo1} are defined by
\begin{shrinkeq}{-1ex}
\begin{align}
\label{equ:normalequ_memo2new} \mathbf{Z}_n &= \mathop{\sum}\limits _{i=1}^{n}\beta^{n-i}\mathbf{y}_{i}\mathbf{x}_i^{\top}\\
\label{equ:normalequ_memo3new} \bm{\Phi}_n &= \mathop{\sum}\limits _{i=1}^{n}\beta^{n-i}\mathbf{x}_i\mathbf{x}_i^{\top} + \delta\beta^{n}\mathbf{I}~.
\end{align}
\end{shrinkeq}
The normal equations in the case of $b = 1$ can be seen as an expansion of the single sample case without summing over all the historical observations:
\begin{shrinkeq}{-1ex}
\begin{align}
\label{equ:normalequ_singlesample} \widehat{\mathbf{W}}_n = \mathbf{y}_{n}\mathbf{x}_n^{\top}\left(\mathbf{x}_n\mathbf{x}_n^{\top} + \delta\mathbf{I}\right)^{-1}~.
\end{align}
\end{shrinkeq}

Isolating the terms corresponding to $i = n$ from the rest of the summations in Eqs.~\eqref{equ:normalequ_memo2new} and \eqref{equ:normalequ_memo3new} yields the following recursions for updating $\mathbf{Z}_n$ and $\bm{\Phi}_n$ respectively:
\begin{shrinkeq}{-1ex}
\begin{align}
\label{equ:normalequ_memo4} \mathbf{Z}_n &= \beta\mathbf{Z}_{n-1} + \mathbf{y}_{n}\mathbf{x}_n^{\top}~,\\
\label{equ:normalequ_memo5} \bm{\Phi}_n &= \beta\bm{\Phi}_{n-1} + \mathbf{x}_n\mathbf{x}_n^{\top}~.
\end{align}
\end{shrinkeq}
With $\bm{\Phi}_n$ assumed to be non-singular and thus invertible, we may obtain the following recursive equation for the inverse of $\bm{\Phi}_n$ by applying the Sherman-Morris formula~\cite{Hager1989Inverse}
\begin{shrinkeq}{-1ex}
\begin{align}
\label{equ:inverse_Phi} \bm{\Phi}_n^{-1} = \beta^{-1}\bm{\Phi}_{n-1}^{-1} - \frac{\beta^{-2}\bm{\Phi}_{n-1}^{-1}\mathbf{x}_n\mathbf{x}_n^{\top}\bm{\Phi}_{n-1}^{-1}}{1 + \beta^{-1}\mathbf{x}_n^{\top}\bm{\Phi}_{n-1}^{-1}\mathbf{x}_n}~.
\end{align}
\end{shrinkeq}
If we denote $\mathbf{P}_{n} = \bm{\Phi}_n^{-1}$ and let
\begin{shrinkeq}{-1ex}
\begin{align}
\label{equ:kn} \mathbf{k}_{n} = \frac{\beta^{-1}\mathbf{x}_n^{\top}\mathbf{P}_{n-1}}{1 + \beta^{-1}\mathbf{x}_n^{\top}\mathbf{P}_{n-1}\mathbf{x}_n}~,
\end{align}
\end{shrinkeq}
then Eq.~\eqref{equ:inverse_Phi} can be rewritten as
\begin{shrinkeq}{-1ex}
\begin{align}
\label{equ:inverse_Phi2} \mathbf{P}_{n} = \beta^{-1}\mathbf{P}_{n-1} - \beta^{-1}\mathbf{P}_{n-1}\mathbf{x}_n\mathbf{k}_{n}~.
\end{align}
\end{shrinkeq}
It is surprising to find that multiplying $\mathbf{x}_n^{\top}$ by each side of Eq.~\eqref{equ:inverse_Phi2} yields the following simple expression 
\begin{shrinkeq}{-1ex}
\begin{align}
\label{equ:kn2} \mathbf{x}_n^{\top}\mathbf{P}_{n} &\!=\! \beta^{-1}\mathbf{x}_n^{\top}\mathbf{P}_{n-1} - \beta^{-1}\mathbf{x}_n^{\top}\mathbf{P}_{n-1}\mathbf{x}_n\mathbf{k}_{n} \!=\! \mathbf{k}_{n}~.
\end{align}
\end{shrinkeq}
This facilitates the derivation for obtaining a recursive formula for the normal equations in the case of $b = 1$ when we substitute Eqs.~\eqref{equ:normalequ_memo4} and \eqref{equ:inverse_Phi2} into Eq.~\eqref{equ:normalequ_memo1}
\begin{shrinkeq}{-1ex}
\begin{align}
\label{equ:normalequ_memo6} \widehat{\mathbf{W}}_n &= \beta\mathbf{Z}_{n-1}\mathbf{P}_{n} + \mathbf{y}_{n}\mathbf{x}_n^{\top}\mathbf{P}_{n}\nonumber\\
&= \mathbf{Z}_{n-1}\mathbf{P}_{n-1} - \mathbf{Z}_{n-1}\mathbf{P}_{n-1}\mathbf{x}_n\mathbf{k}_{n} + \mathbf{y}_{n}\mathbf{k}_{n}\nonumber\\
&= \widehat{\mathbf{W}}_{n-1} + \left(\mathbf{y}_{n} - \widehat{\mathbf{W}}_{n-1}\mathbf{x}_n\right)\mathbf{k}_{n}\nonumber\\
&= \widehat{\mathbf{W}}_{n-1} - \left(\widehat{\mathbf{W}}_{n-1}\mathbf{x}_n - \mathbf{y}_{n}\right)\mathbf{x}_n^{\top}\mathbf{P}_{n}~,
\end{align}
\end{shrinkeq}
where the expression inside the brackets on the right-hand side of the last line represents the {\it residual} based on the old least-squares estimate of the parameters to be learned. 

Extending the above derivation to the case of $b > 1$ is not straightforward, because the equations of Eqs.~\eqref{equ:normalequ_memo4} and \eqref{equ:normalequ_memo5} will be expanded to
\begin{shrinkeq}{-1ex}
\begin{gather}
\label{equ:normalequ_memo4new} \mathbf{Z}_n = \beta\mathbf{Z}_{n-1} + \bm{Y}_n^{\top}\bm{X}_n~,\\
\label{equ:normalequ_memo5new} \bm{\Phi}_n = \beta\bm{\Phi}_{n-1} + \bm{X}_n^{\top}\bm{X}_n~,
\end{gather}
\end{shrinkeq}
where
\begin{shrinkeq}{-1ex}
\begin{align}
\label{equ:XYmean} \bm{X}_i = \begin{pmatrix}
\mathbf{x}_{i,1}^{\top} \\
\vdots \\
\mathbf{x}_{i,b}^{\top}\\
\end{pmatrix}  ~~~\mathrm{and}~~~  \bm{Y}_i = \begin{pmatrix}
\mathbf{y}_{i,1}^{\top} \\
\vdots \\
\mathbf{y}_{i,b}^{\top}\\
\end{pmatrix}~.
\end{align}
\end{shrinkeq}
This obviously prohibits the derivation of Eq.~\eqref{equ:inverse_Phi}. Alternatively, using the fact that the mean squared error inside the brackets of Eq.~\eqref{equ:costfunctionLSnew2}, which is based on the $l_2$ norm of the {\it residual} vectors for all of the sample pairs in the $i_{th}$ {\it block}, serves as an upper bound of the error term for the virtual sample pair $\left\{\overline{\mathbf{x}}_i, \overline{\mathbf{y}}_i\right\}$, \ie,
\begin{shrinkeq}{-1ex}
\begin{align}
\label{equ:submultiplicativity} \left\Vert \overline{\mathbf{y}}_i - \mathbf{W}_n\overline{\mathbf{x}}_i\right\Vert^2 \leq \frac{1}{b}\left\Vert \bm{Y}_i - \bm{X}_i\mathbf{W}_n^{\top}\right\Vert^2~,
\end{align}
\end{shrinkeq}
where
\begin{shrinkeq}{-1ex}
\begin{align}
\label{equ:mean1} \overline{\mathbf{x}}_i = \frac{1}{b}\mathop{\sum}\limits _{j=1}^{b}\mathbf{x}_{i, j} ~~~\mathrm{and}~~~ \overline{\mathbf{y}}_i = \frac{1}{b}\mathop{\sum}\limits _{j=1}^{b}\mathbf{y}_{i, j}~,
\end{align}
\end{shrinkeq} 
we can instead approximately obtain the recursive formula for the normal equations in the case of $b > 1$ based on this virtual sample pair:
\begin{shrinkeq}{-1ex}
\begin{align}
\label{equ:RLS_update1_2} \widehat{\mathbf{W}}_n = \widehat{\mathbf{W}}_{n-1} - \left(\widehat{\mathbf{W}}_{n-1}\overline{\mathbf{x}}_n - \overline{\mathbf{y}}_{n}\right)\overline{\mathbf{x}}_n^{\top}\mathbf{P}_{n}~,
\end{align}
\end{shrinkeq}
where $\mathbf{P}_{n}$ is updated based on $\overline{\mathbf{x}}_n$ using Eqs.~\eqref{equ:kn} and \eqref{equ:inverse_Phi2}.

In practice, the exact optimum value $\widehat{\mathbf{W}}_{n-1}$ is hardly to be obtained and its approximate optimum value $\widehat{\mathbf{W}}_{n-1}^{appr}$ is commonly obtained from some optimization algorithm. However, we cannot directly use Eq.~\eqref{equ:RLS_update1_2} to obtain $\widehat{\mathbf{W}}_{n}^{appr}$ as the relationship between $\widehat{\mathbf{W}}_{n}^{appr}$ and $\widehat{\mathbf{W}}_{n-1}^{appr}$ does not rigorously comply with Eq.~\eqref{equ:RLS_update1_2}. Inspired from the gradient descent paradigm, we can, for instance, model the changing between $\widehat{\mathbf{W}}_{n-1}^{appr}$ and $\widehat{\mathbf{W}}_{n}^{appr}$ by the following iterations:
\begin{shrinkeq}{-1ex}
\begin{align}
\label{equ:derivativenew} \mathbf{W}_n^{iter+1} &= \mathbf{W}_n^{iter} - \eta \left(\mathbf{W}_n^{iter}\overline{\mathbf{x}}_n - \overline{\mathbf{y}}_{n}\right)\overline{\mathbf{x}}_n^{\top}\mathbf{P}_{n}~,\\
\label{equ:updatenew} \mathbf{W}_n^{0} &= \widehat{\mathbf{W}}_{n-1}^{appr}~,
\end{align}
\end{shrinkeq}
where $\eta$ is another pre-defined learning rate. From the above equations, we see that the approximate optimum value $\widehat{\mathbf{W}}_{n}^{appr}$ is obtained only based on the most recently arrived data {\it block} $n$, though it is dedicated to satisfying the {\it cost function} in Eq.~\eqref{equ:costfunctionLSnew2} and hence retaining the historical memory. Consequently using the fact that the matrix product $\left(\mathbf{W}_{n}^{iter}\overline{\mathbf{x}}_n - \overline{\mathbf{y}}_{n}\right)\overline{\mathbf{x}}_n^{\top}$ represents the gradient with respect to the weighting parameters $\mathbf{W}_{n}^{iter}$ for the virtual input $\overline{\mathbf{x}}_n$ of the $n_{th}$ data {\it block} when considering its squared-error loss $\frac{1}{2}\left\Vert \overline{\mathbf{y}}_n - \mathbf{W}_n^{iter}\overline{\mathbf{x}}_n\right\Vert^2$ without the $l_2$ regularization term, we can cast the iterations in Eqs.~\eqref{equ:derivativenew} and \eqref{equ:updatenew} as an improved gradient descent algorithm for online learning with memory retention. Note that the initial condition of $\mathbf{P}_{n}$ is set to
\begin{shrinkeq}{-1ex}
\begin{align}
\label{equ:initialcondition} \mathbf{P}_{0} = \bm{\Phi}_0^{-1} = \left(\delta\mathbf{I}\right)^{-1} = \left.\mathbf{I}\middle/\delta\right.~.
\end{align}
\end{shrinkeq}

\textit{Discussion.} The traditional gradient descent algorithms following the {\it sliding-window} strategy for online learning compute the gradient at each optimization iteration based on the {\it cost function} similar to Eq.~\eqref{equ:costfunctionLS} for the whole batch of fixed-window data (BGD) or its mini-batch after random shuffle operation (MBSGD). An alternative strategy for improving these traditional algorithms to retain memory by performing minimal changes of their settings and implementations is to cast the whole batch (for BGD) or one mini-batch (for MBSGD) data as a data {\it block} at some time index in Eq.~\eqref{equ:costfunctionLSnew2}, and then improve the optimization iterations based on our proposed approach in Eqs.~\eqref{equ:derivativenew} and \eqref{equ:updatenew}. Note that the gradient term based on the virtual sample pair in Eq.~\eqref{equ:derivativenew} can be approximately replaced with the real gradient at each optimization iteration computed from BGD or MBSGD.

\section{Recursive Formulation for Multiple Fully-connected Layers in RT-MDNet}
\label{Sec:MLP}
In the online deep tracker RT-MDNet~\cite{Jung2018RTMDNet} that is dedicated to target classification, the multi-layer perceptron (MLP) with several fully-connected $1$-$D$ layers is applied to the last stage of the deep learning architecture for classification. The features present in the final $2$-$D$ feature maps are concatenated into one long input vector to the following MLP. In this section, we will show how our proposed RLS estimator-aided online learning approach for memory retention can be realized in this single-head real-time tracker to online learn those MLP layers.

\textit{Online Learning in RT-MDNet.} Being identical to MDNet~\cite{Nam2016MDNet}, RT-MDNet only updates the fully-connected MLP layers (\verb'fc4-6') $\left\{\mathbf{W}^l\right\}_{l=4}^6$ in an online manner while keeping the convolutional feature extraction layers (\verb'conv1-3') $\left\{\mathbf{W}^l\right\}_{l=1}^3$ fixed, namely update stage. Before that, the MLP layers also need to be fine-tuned using the samples from the initial frame to customize themselves to a new testing domain, namely initialization stage. There are two types of memory being maintained in the update stage to make compromises between robustness and adaptiveness: the long-term memory for regular updates with the samples collected for a long period of time; the recent short-term memory with the occasional updates triggered whenever the score of the estimated target is below a threshold, indicating the unreliable tracking.

During online tracking, the cross-entropy loss based on the softmax output of the single fully-connected binary classification layer $\mathbf{W}^6$ is used to fine-tune or update all the MLP layers and the optimization algorithm of MBSGD is exploited, which is a compromise between BGD and the single instance SGD. Specifically, suppose there are $b$ samples in one mini-batch used for one optimization iteration of the MBSGD process, it is thus the following partial derivatives over the mean value of the Binary Cross Entropy (BCE) loss for all the $b$ samples that is used for updating $\mathbf{W}^l$ of each layer:
\begin{shrinkeq}{-1ex}
\begin{align}
\label{equ:derivative} \Delta \mathbf{W}^l &\!=\! \nabla _{\mathbf{W}^l}\left[\frac{1}{b}\mathop{\sum}\limits _{j=1}^{b}BCE\left(\mathbf{x}_j, \mathbf{y}_{j}, \left\{\mathbf{W}^l\right\}_{l=4}^6\right)\right] \!+\! \lambda^l_{r}\mathbf{W}^l~,\\
\label{equ:update} \mathbf{W}^l &\longleftarrow \mathbf{W}^l - \eta^l_{r} \Delta \mathbf{W}^l~,
\end{align}
\end{shrinkeq}
where $\mathbf{x}_j$ is the long input vector of the sample $j$ in the current batch, $\mathbf{y}_j$ is the class (positive/negative) prediction vector, and $\lambda^l_{r}$ and $\eta^l_r$ are the weight decay and pre-defined learning rate respectively in the MBSGD process of RT-MDNet. The weight decay encourages the learned weights to be small in magnitude to improve the generalization performance of them~\cite{Zhang2019WeightDecay}.

\begin{algorithm}[!t]
\setstretch{1.1}
\small
\DontPrintSemicolon
  \KwIn{Off-the-shelf deep RT-MDNet tracking model with multiple layers $\left\{\mathbf{W}^l\right\}_{l=1}^6$, initial $\mathbf{P}^l_0 = \left.\mathbf{I}\middle/\delta^l_r\right.$ for each fully-connected layer ($4\leq l \leq 6$), and test sequence with first frame annotated}
  \KwOut{Estimated target states in the rest frames}
  \lnl{InRes1}Pre-process identical to RT-MDNet, \eg, draw positive sample set $S_1^+$ and negative sample set $S_1^-$\;
  \lnl{InRes2}Fine-tune $\left\{\mathbf{W}^l\right\}_{l=4}^6$ using $S_1^+ \!\cup\! S_1^-$ in initialization stage:\;
  \For{the $n_{\mathrm{th}}$ improved MBSGD iteration}{
  Update $\mathbf{P}^l_n$ based on the virtual input $\overline{\mathbf{x}}_n^l$ using Eqs.~\eqref{equ:kn} and \eqref{equ:inverse_Phi2}\;
  Update $\mathbf{W}^l \leftarrow \mathbf{W}^l_n$ using Eqs.~\eqref{equ:derivative} and \eqref{equ:RLS_update_RTMDNet}\;
  }
  \lnl{InRes3}Initialize the frame index set $\mathcal{T} \leftarrow \{1\}$\;
  \Repeat{end of sequence}{
      \lnl{InRes4}Find the optimal target state like in RT-MDNet\;
      \If{\texttt{target score} $> 0$}{\lnl{InRes5}Draw $S_t^+$ and $S_t^-$, and set $\mathcal{T} \leftarrow \mathcal{T} \cup \{t\}$\;
	  \lnl{InRes6}\lIf{$|\mathcal{T}| > \tau$}{$\mathcal{T} \leftarrow \mathcal{T} \setminus \left\{\min_{\upsilon \in \mathcal{T}}\upsilon\right\}$}      
      }
      \If{\texttt{target score} $\leq 0$}{\lnl{InRes7}\lIf{$\left\{\mathbf{W}^l_{\mathrm{bk}}\right\}_{l=4}^6 = \varnothing$}{create backups $\left\{\mathbf{W}^l_{\mathrm{bk}}\right\}_{l=4}^6 \leftarrow \left\{\mathbf{W}^l\right\}_{l=4}^6$}
	  \lnl{InRes8}Occasionally update $\left\{\mathbf{W}^l\right\}_{l=4}^6$ using $S_{\upsilon \in \mathcal{T}}^+ \cup S_{\upsilon \in \mathcal{T}}^-$ based on the original MBSGD
      }
      \ElseIf{$t$ \textbf{\texttt{mod}} $10 = 0$}{\lnl{InRes9}\lIf{$\left\{\mathbf{W}^l_{\mathrm{bk}}\right\}_{l=4}^6\! \neq \!\varnothing$}{recover backups $\left\{\mathbf{W}^l\right\}_{l=4}^6 \!\! \leftarrow \!\! \left\{\mathbf{W}^l_{\mathrm{bk}}\right\}_{l=4}^6$, and set $\left\{\mathbf{W}^l_{\mathrm{bk}}\right\}_{l=4}^6 = \varnothing$}
      \lnl{InRes10}Regularly update $\left\{\mathbf{W}^l\right\}_{l=4}^6$ using $S_{\upsilon \in \mathcal{T}}^+ \cup S_{\upsilon \in \mathcal{T}}^-$:\;
      \For{the $n_{\mathrm{th}}$ improved MBSGD iteration}{
  Update $\mathbf{P}^l_n$ based on the virtual input $\overline{\mathbf{x}}_n^l$ using Eqs.~\eqref{equ:kn} and \eqref{equ:inverse_Phi2}\; 
  Update $\mathbf{W}^l \leftarrow \mathbf{W}^l_n$ using Eqs.~\eqref{equ:derivative} and \eqref{equ:RLS_update_RTMDNet}\;
  }
      }
      }
\caption{RLS Estimator-Aided RT-MDNet}
\label{algo:RLS_RTMDNet}
\end{algorithm}

\textit{RLS Estimator-Aided RT-MDNet.} In our preliminary version~\cite{Gao2020RLS-RTMDNet}, we have claimed and demonstrated that each layer of MLP in the case of squared-error loss function can be represented by the system of normal equations for RLS estimation by simply ignoring non-linear activation functions. In the present work, we add more explanations on how to extend the derivation of normal equations for MLP in~\cite{Gao2020RLS-RTMDNet} to generally handle other layers including non-linear activation functions and leave them in the supplementary material due to space constraints. We also provide more explanations on why all the derivations in the demonstration can be easily transplanted to the case of cross-entropy loss in this supplementary material and refer the readers to~\cite{Saitoh2021DLB} for more details. This means we can exploit our proposed RLS estimator-aided online learning approach to improve the updating of each MLP layer in RT-MDNet in order to retain memory. In specific, if we cast one mini-batch of samples used for one optimization iteration of the MBSGD process as a data {\it block} at some time index $n$ as in Section~\ref{Sec:RLS}, then this optimization iteration can be carried out as follows
\begin{shrinkeq}{-1ex}
\begin{align}
\label{equ:RLS_update_RTMDNet} \mathbf{W}_n^{l} &= \mathbf{W}_{n-1}^{l} - \widetilde{\eta}^l_{r}  \Delta \mathbf{W}_n^{l}\mathbf{P}_{n}^l~,\\
\label{equ:RLS_update2_RTMDNet} \mathbf{P}_{0}^l &= \left.\mathbf{I}\middle/\delta^l_r\right.~,
\end{align}
\end{shrinkeq}
for memory retention, where $\widetilde{\eta}^l_{r}$ is the new learning rate, $\delta^l_r$ is the positive constant for RLS estimation and the term $\mathbf{W}_{n-1}^l\left(\mathbf{I} - \widetilde{\eta}^l_{r}\lambda^l_{r}\mathbf{P}_{n}^l\right)$ serves as the weight decay for optimization. Note that Eq.~\eqref{equ:RLS_update_RTMDNet} is only implemented once for improving the MBSGD process, in contrast to Eq.~\eqref{equ:derivativenew}, as the same mini-batch of samples may appear several times for optimization at each update step. $\mathbf{P}_{n}^l$ for each layer is updated using Eqs.~\eqref{equ:kn} and \eqref{equ:inverse_Phi2} based on each layer's virtual input
\begin{shrinkeq}{-1ex}
\begin{align}
\label{equ:mean2} \overline{\mathbf{x}}_n^l = \frac{1}{b}\mathop{\sum}\limits _{j=1}^{b}\mathbf{x}_{j}^l~,
\end{align}
\end{shrinkeq}
where $\mathbf{x}_{j}^l$ is the input to layer $l$ for the sample $j$.

Since we only concentrate on the online tracking part to improve RT-MDNet, we leave all the other parts of RT-MDNet almost unchanged including, for example, the improved RoIAlign technique, all their hyper-parameter counterparts and the off-the-shelf model trained with additional instance embedding loss in their paper. As the RLS estimator-aided online learning can overcome catastrophic forgetting of historical memory, we thus do not need maintaining the long-term memory to achieve robustness any more. That is to say, we still use the same recent short-term memory for RLS estimator-aided regular updates, while the setting of the occasional updates triggered by unreliable tracking is preserved intact, in that there may be failure cases for the model based on the regular updates and the model with the occasional updates customized (overfitting) to the recent short-term memory may work well with more discrimination. Note that the regularly updated model is fixed during the occasional updates. More details about the improved RT-MDNet with the RLS estimator-aided online learning, namely RLS-RTMDNet, are explained in Algorithm \ref{algo:RLS_RTMDNet}.

\section{Recursive Formulation for Convolutional Layers in DiMP}
\label{Sec:CNN}
In the recent state-of-the-art online tracker DiMP~\cite{Bhat2019DiMP} that is dedicated to both target classification and location estimation, high level knowledge is incorporated to train the target location estimation component, following its pioneering work ATOM~\cite{Danelljan2019ATOM}, through extensive offline
learning so as to predict the overlap between the target object and an estimated
bounding box; meanwhile a more powerful classification component is carefully designed to incorporate additional prior knowledge for fast initial adaptation through the offline meta-training process and finally online updated to guarantee high discriminative power in the presence of distractors. In specific, the target classification component has its target model constitute the filter weights of a convolutional layer and provides target classification scores as regression output without non-linear output activation. It  also employs the block features from a ResNet-based backbone, which is pre-trained on ImageNet~\cite{Russakovsky2015ImageNet} and fine-tuned on tracking datasets along with the target classification and location estimation components. In this section, we will show how our proposed RLS estimator-aided online learning approach for memory retention can be realized in this multi-head real-time tracker to online learn the convolutional layer.

\textit{Online Learning in DiMP.} Besides the target model's filter weights, the target classification component also includes an initializer network for initial filter prediction and several free parameter predictors for setting the target mask, spatial weight, regularization factor and even regression target in the discriminative learning loss. In the first frame, the filter weights are initialized based on the initializer network and then fine-tuned based on the initial training samples from data augmentation, namely initialization stage. Subsequently, the extracted feature map annotated by a Gaussian centered at the estimated target location in every frame is added to the fixed-window sample set as a training sample. During online tracking, the convolutional layer with a single output channel is online optimized every $20_{th}$ frame, namely update stage. 

Given the training set of samples, we can approximately formulate the online learning objective in DiMP as follows based on the $l_2$ norm of the {\it residuals}:
\begin{shrinkeq}{-1ex}
\begin{align}
\label{equ:costfunctionLSATOM} \mathscr{L}(\mathbf{W}) = \mathop{\sum}\limits _{j=1}^{N}\left\Vert \bm{\Gamma}_j \odot \left(\mathbf{Y}_j - \mathbf{W}\ast\mathbf{X}_j\right)\right\Vert^2 + \frac{\lambda_d}{2}\left\Vert \mathbf{W}\right\Vert^2~.
\end{align}
\end{shrinkeq}
Here, $\mathbf{X}_j$ is the extracted feature map of the sample $j$, $\mathbf{Y}_j$ is its annotated Gaussian confidence map, $\mathbf{W}$ represents the weights of the convolutional layer, $\ast$ denotes standard multi-channel convolution, $\lambda_d$ is the weight decay parameter, $N$ is the fixed-window size, $\odot$ denotes Hadamard product, and the impact of each training sample along with its spatial weight is controlled by the weighting factor $\bm{\Gamma}_j$.

\begin{algorithm}[!t]
\setstretch{1.1}
\small
\DontPrintSemicolon
  \KwIn{Off-the-shelf deep DiMP tracking model with one convolutional layer $\mathbf{W}$ for target classification, initial $\mathbf{P}_0 = \left.\mathbf{I}\middle/\delta_d\right.$ for this layer $\mathbf{W}$, and test sequence with first frame annotated}
  \KwOut{Estimated target states in the rest frames}
  \lnl{InRes1}Initialize the fixed-window sample set $S$ with the data augmented samples extracted from the first frame\;
  \lnl{InRes2}Predict a reasonable initial estimate of $\mathbf{W}$ and then fine-tune it using the SD methodology identical to DiMP in initialization stage, and set $\widehat{\mathbf{W}}_{0}^{appr} = \mathbf{W}$ in the following\;
  \Repeat{end of sequence}{
      \lnl{InRes3}Find the optimal target state like in DiMP\;
      \If{\texttt{update\_flag}}{\lnl{InRes4}Update the sample set $S$ along with its sample weights identical to DiMP\;
      }
      \If{$t-1$ \textbf{\texttt{mod}} $20 = 0$ or \texttt{hard\_negative}}{
      \For{the $n_{\mathrm{th}}$ improved BGD procedure}{
      Set $\mathbf{W}_n^{0} = \widehat{\mathbf{W}}_{n-1}^{appr}$\;
      Update $\mathbf{P}_n$ based on the virtual input $\overline{\mathbf{x}}_n$ using Eqs.~\eqref{equ:kn} and \eqref{equ:inverse_Phi2}\; 
      \Repeat{end of iteration}{
      Compute the current gradient $\Delta \mathbf{W}_n^{iter}$\;
      Update $\mathbf{W}_n^{iter}$ based on Eq.~\eqref{equ:derivativenewATOM}\;
      }
  	  Update $\mathbf{W} \leftarrow \widehat{\mathbf{W}}_{n}^{appr}$\; 
  }
      }
      }
\caption{Recursive LSE-Aided DiMP}
\label{algo:RLS_DiMP}
\end{algorithm}

In modern deep learning architectures, each convolutional layer is usually converted to a general matrix-matrix multiplication (GEMM) operation for fast execution on modern CPUs or GPUs. This is achieved by an unfolding and duplication of the input and an unrolling and concatenation of the kernels to produce the input matrix and kernel matrix. These transformations are realized based mainly on a function $$f(\cdot)=img2col(\cdot)$$ in the modern deep learning libraries. Consequently, the $l_2$ norm term in Eq.~\eqref{equ:costfunctionLSATOM} can be rewritten as
\begin{shrinkeq}{-1ex}
\begin{align}
\label{equ:costfunctionLSATOM2} &\left\Vert \bm{\Gamma}_j \odot\left(\mathbf{Y}_j - \mathbf{W}\ast\mathbf{X}_j\right)\right\Vert^2\nonumber\\=& \left\Vert f^{\top}\left(\bm{\Gamma}_j\right) \odot \left(f^{\top}\left(\mathbf{Y}_j\right) - f^{\top}\left(\mathbf{W}\right)f\left(\mathbf{X}_j\right)\right)\right\Vert^2~.
\end{align}
\end{shrinkeq}
If we denote the column number of $f\left(\mathbf{X}_j\right)$ as $M$, the $k_{th}$ column in $f\left(\mathbf{X}_j\right)$ as $\mathbf{x}_{jk}$, the $k_{th}$ element in $f\left(\mathbf{Y}_j\right)$ as $y_{jk}$ and the $k_{th}$ element in $f\left(\bm{\Gamma}_j\right)$ as $\gamma_{jk}$, then the learning objective of Eq.~\eqref{equ:costfunctionLSATOM} can be rewritten as
\begin{shrinkeq}{-1ex}
\begin{align}
\label{equ:costfunctionLSATOM3} \mathscr{L}(\mathbf{W}) \!=\! \mathop{\sum}\limits _{j=1}^{N}\mathop{\sum}\limits _{k=1}^{M}\left\Vert \sqrt{\gamma_{jk}}\left(y_{jk} \!-\! f^{\top}\left(\mathbf{W}\right)\mathbf{x}_{jk}\right)\right\Vert^2 \!+\! \frac{\lambda_d}{2}\left\Vert \mathbf{W}\right\Vert^2\!.
\end{align}
\end{shrinkeq}

During online tracking, the BGD optimization or the SD methodology can be used to update $\mathbf{W}$ based on the above loss function. Specifically, in the case of BGD optimization, it is the following partial derivatives over the above squared-error loss for all the $N$ samples that is used for updating $\mathbf{W}$ at each optimization iteration of the BGD process:
\begin{shrinkeq}{-1ex}
\begin{align}
\label{equ:derivativeATOM} \Delta \mathbf{W} &= \nabla _{\mathbf{W}}\mathscr{L}(\mathbf{W})~,\\
\label{equ:updateATOM} \mathbf{W} &\longleftarrow \mathbf{W} - \eta_d \Delta \mathbf{W}~,
\end{align}
\end{shrinkeq}
where $\eta_d$ is a pre-defined learning rate.

\textit{RLS Estimator-Aided DiMP.} From Eq.~\eqref{equ:costfunctionLSATOM3} we can easily find that our proposed RLS estimator-aided online learning approach is also able to improve the updating of the convolutional layer $\mathbf{W}$ in DiMP in order to retain memory. This is achieved by applying BGD optimization method in the update stage of DiMP and then improving the updating process. In specific, if we cast all columns of $f\left(\mathbf{X}_j\right)$ for all the training samples in the $N$-sized sample set as a data {\it block} at some time index $n$ as in Section~\ref{Sec:RLS}, then the optimization of our improved BGD process can be carried out as follows
\begin{shrinkeq}{-1ex}
\begin{align}
\label{equ:derivativenewATOM} f\left(\mathbf{W}_n^{iter+1}\right) &= f\left(\mathbf{W}_n^{iter}\right) - \widetilde{\eta}_d f\left(\Delta \mathbf{W}_n^{iter+1}\right) \mathbf{P}_{n}~,\\
\label{equ:updatenewATOM} \mathbf{W}_n^{0} &= \widehat{\mathbf{W}}_{n-1}^{appr}~,\\
\label{equ:updatenewATOM2} \mathbf{P}_{0} &= \left.\mathbf{I}\middle/\delta_d\right.~,
\end{align}
\end{shrinkeq}
where $\widehat{\mathbf{W}}_{n-1}^{appr}$ is the last approximate optimum value, $\widetilde{\eta}_d$ is the new learning rate, $\delta_d$ is the positive constant for RLS estimation, the term $f\left(\mathbf{W}_n^{iter}\right)\left(\mathbf{I} - \widetilde{\eta}_d\lambda_d\mathbf{P}_{n}\right)$ serves as the weight decay for optimization, and $\mathbf{P}_{n}$ is updated using Eqs.~\eqref{equ:kn} and \eqref{equ:inverse_Phi2} based on the virtual input, which is the mean value of the data {\it block} at the current time index $n$:
\begin{shrinkeq}{-1ex}
\begin{align}
\label{equ:mean3} \overline{\mathbf{x}}_n =\frac{1}{\sqrt{NM}} \mathop{\sum}\limits _{j=1}^{N}\mathop{\sum}\limits _{k=1}^{M}\sqrt{\gamma_{jk}}\mathbf{x}_{jk}.
\end{align}
\end{shrinkeq}

Since we only concentrate on the online tracking part to improve DiMP, we also leave all the other parts of DiMP almost unchanged including, for example, the default off-the-shelf model trained using the train splits of LaSOT, TrackingNet, GOT10k and MS-COCO, all their hyper-parameter counterparts and the settings of initialization stage to achieve $\widehat{\mathbf{W}}_{0}^{appr}$. It is noteworthy that the optimization strategy of the SD methodology is preserved intact along with all its parameter settings in the initialization stage. The reason is that the SD methodology exhibits superior convergence speed compared to the BGD optimization strategy as claimed in~\cite{Bhat2019DiMP}. More details about the improved DiMP with the RLS estimator-aided online learning, namely RLS-DiMP, are explained in Algorithm \ref{algo:RLS_DiMP}.

\section{Experiments}
\label{Sec:EXP}
\subsection{Implementation Details and Baseline Setup}
\label{Section:setup}

We here only show the hyper-parameters specific to our recursive LSE-aided online learning, and leave all the other hyper-parameters and the off-the-shelf tracking models shared with the original RT-MDNet and DiMP trackers unchanged\footnote {We refer the reader to~\cite{Jung2018RTMDNet} and \cite{Bhat2019DiMP} for their parameter settings and their default off-the-shelf tracking models can be achieved at \texttt{\url{https://github.com/IlchaeJung/RT-MDNet}} and \texttt{\url{https://github.com/visionml/pytracking/blob/master/MODEL_ZOO.md\#Models}} respectively.} unless otherwise specified. 

In our previous implementation of  RLS-RTMDNet in~\cite{Gao2020RLS-RTMDNet}, $\delta^l_r$ in Algorithm~\ref{algo:RLS_RTMDNet} and the parameter to measure the memory in Eq.~\eqref{equ:normalequ_memo3new} for each of the fully-connected layers $\mathbf{W}^4$, $\mathbf{W}^5$ and $\mathbf{W}^6$ were simply set to $1.0$; the learning rates $\widetilde{\eta}^l_{r}$ for them in Eq.~\eqref{equ:RLS_update_RTMDNet} in both initialization and update stages were fixed to $0.01$, $0.01$ and $0.1$. According to~\cite{Haykin2002AFT, Moustakides1997}, however, initializing $\bm{\Phi}_0$ with a ``small" value for the most application scenarios will help the recursive LSE-aided online learning to achieve its optimal performance, which is insensitive to the exact ``small" value used. So we experimentally found some new configurations for RLS-RTMDNet in this paper, and the experimental results show that roughly setting $\delta^l_r$ of each layer to $5e^{-4}$ and directly inheriting the learning rates $\widetilde{\eta}^l_{r}$ from the original RT-MDNet tracker give rise to a further performance improvement over our previous tracking results in~\cite{Gao2020RLS-RTMDNet}, especially with a significant improvement on \textit{UAV123}~\cite{Mueller2016UAV} . We re-implemented the public source code of the original RT-MDNet as a baseline and followed all of its default settings to demonstrate the effectiveness of our proposed online learning framework in the case of the MLP layers. Although we leave many settings of RT-MDNet unchanged in our RLS-RTMDNet tracker, there are still some differences in the online tracking part related to the solely maintained short-term memory and the weight backups, in addition to the improved MBSGD. So we prepared another baseline by only replacing the improved MBSGD with original MBSGD in Algorithm \ref{algo:RLS_RTMDNet}, leading to a degraded version, namely simpleRT-MDNet.

We also selected DiMP to demonstrate the effectiveness of our proposed online learning framework in the case of the convolutional layer. For the updating of $\mathbf{W}$ in Algorithm~\ref{algo:RLS_DiMP}, we simply set $\delta_d$ to $0.1$ and the parameter to measure the memory in Eq.~\eqref{equ:normalequ_memo3new} to $1.0$. The feature dimensionality of $\mathbf{X}_j$ in Eq.~\eqref{equ:costfunctionLSATOM} is $512$, which makes the dimension size of $\mathbf{P}_{n}$ large enough for memory retention. Besides re-implementing the public source code of DiMP as one baseline, we add three more kinds of baselines for a more in-depth analysis. First, replacing only the improved BGD with vanilla BGD in Algorithm~\ref{algo:RLS_DiMP} leads to the baseline termed as BGD-DiMP, which can be used to directly validate the effectiveness of our approach. We experimentally found the optimal learning rates for BGD-DiMP while performing the BGD optimization with five recursions every $20_{th}$ frame, \ie, $3e^{-2}$ for short-term tracking benchmarks (the average sequence length does not exceed 600 frames), and $3e^{-3}$ for long-term ones. This learning rate setting is also applied in our RLS-DiMP for fair comparison. Second, increasing the SD optimizer recursion number from two to five for updating $\mathbf{W}$ every $20_{th}$ frame in the original DiMP leads to the baseline denoted by DiMP$^*$, which can be used to study the  overfitting issue caused by fast converging to a {\it local} point with respect to the fixed-sized sample set at each update step. Finally, considering that the exponential moving average (EMA) strategy is widely applied in a slow-moving fashion for striking a compromise between adapting quickly and maintaining robustness in visual tracking~\cite{Bolme2010MOSSE, Henriques2015KCF, Danelljan2015SRDCF, Danelljan2016CCOT, Bertinetto2016Staple, Danelljan2017DSST, Bertinetto2017CFNet, Galoogahi2017BACF, YLi2019ARCF, Xu2019LADCF, Xu2019GFS-DCF, YLi2020AMCF, Wang2019UDT}, or stabilizing the network representation with smoother changes in self-supervised learning~\cite{meanteacher, mocov1, BYOL}, we can also use this strategy to combine the newly updated parameters with previous ones to save historical information and avoid overfitting. By setting different EMA coefficient for discounting older parameters in slow-moving, moderate-moving and fast-moving ways, we can achieve a new kind of baselines for comparison (see Sec.~\ref{sec:Ablation_RLS-DiMP} for more details). Our RLS-DiMP and all the above baselines are implemented with a ResNet50 backbone~\cite{He2016ResNet}.

Note that all the evaluations of our methods and the re-implemented baselines are conducted by running them several times due to the stochastic nature of RT-MDNet and DiMP, and performed on a single Nvidia RTX 3090 GPU accompanied by an Intel(R) Xeon(R) Gold 6248R CPU @ 3.00GHz.

\subsection{Experimental Analyses of RLS-RTMDNet}
\label{Section:experimentRLS-RTMDNet}

We evaluate the proposed RLS estimator-aided online RLS-RTMDNet thoroughly over \textit{OTB-2015}~\cite{Wu2015OTB}, \textit{UAV123}~\cite{Mueller2016UAV} and \textit{VOT2016/2017}~\cite{Kristan2016VOTconf, Kristan2017VOTconf} benchmarks by following rigorously the evaluation protocols. We choose these small-scale tracking benchmarks for this evaluation due to the reason that the off-the-shelf RT-MDNet tracking model is trained without additionally relying on the \textit{MS-COCO} dataset~\cite{Lin2014COCO} or recently proposed large-scale tracking datasets, which makes it hard to achieve good results on the more challenging large-scale tracking benchmarks and thus do meaningful analyses.

\subsubsection{Ablation study}
We start by summarizing the ablation study comparison of our proposed RLS-RTMDNet with its corresponding baselines on the \textit{OTB-2015}  and \textit{UAV123} benchmarks in Table~\ref{table:ablationstudy_RLS_RTMDNet}. \textit{OTB-2015} includes 100 objects to be tracked in 98 challenging sequences, and provides a no-reset One-Pass Evaluation (OPE) criterion by running test trackers until the end of a sequence. In its success plot for quantifying performance, the success rate ($SR$) refers to the mean overlap precision ($OP$) score over all sequences and is plotted against a uniform range of some thresholds between 0 and 1. An area-under-the-curve ($AUC$) metric can also be computed to rank the trackers. The $OP$ score is the fraction of frames in a sequence where the inter-section-over-union overlap of the predicted and ground-truth rectangles exceeds a given threshold. In its precision plot, the precision encodes the proportion of frames where the Euclidean distance between the predicted and ground-truth centroids is smaller than a given threshold. \textit{UAV123} compiles a larger set of 123 videos (with more than $110K$ frames) captured by low-altitude UAVs, which is inherently different from \textit{OTB-2015} despite that it also uses the same evaluation protocol to \textit{OTB-2015}. 

\begin{table}
\small
\setlength{\tabcolsep}{2.6pt} \small
    \fontsize{8pt}{10pt}\selectfont
\centering
\caption{Ablation Study Comparison of RLS-RTMDNet with Its Corresponding Baselines RT-MDNet and simpleRT-MDNet on the \textit{OTB-2015}  and \textit{UAV123} Benchmarks. $Prec.$ and $Succ.$ Denote Precision Score at 20 Pixels and $AUC$ of Success Plot Respectively. The Average Over 50 Runs Is Reported for These MDNet-Style Trackers.}
\begin{tabular}{c!{\vrule}cc!{\vrule}cc!{\vrule}c!{\vrule}c}
\toprule
\multirow{2}{*}{Trackers} & \multicolumn{2}{c!{\vrule}}{\textit{OTB-2015}} & \multicolumn{2}{c!{\vrule}}{\textit{UAV123}} & Running & GPU Memory\\
\cmidrule(lr){2-5}
&  $Prec.$ &  $Succ.$ &  $Prec.$ &   $Succ.$  & Speed & Consumption \\
\midrule
RT-MDNet & 0.857  & 0.634 & 0.723 & 0.513 & $\sim$32 \emph{fps} & $<$2400MiB \\
\midrule
simpleRT-MDNet & 0.854 & 0.630 & 0.722  & 0.509 & $\sim$32 \emph{fps} & $<$2400MiB  \\
\textbf{RLS-RTMDNet} & 0.871  & 0.644 & 0.727  & 0.525 & $\sim$31 \emph{fps} & $<$2400MiB \\
\end{tabular}
\label{table:ablationstudy_RLS_RTMDNet}
\end{table}

It is clearly and statistically shown in Table~\ref{table:ablationstudy_RLS_RTMDNet} that our online learning approach can unveil the power of RT-MDNet without incurring too much extra computational and memory cost. Specifically, without bells and whistles, our RLS-RTMDNet improves all the precision (at 20 pixels) and $AUC$ scores over the baseline RT-MDNet on both the \textit{OTB-2015} and \textit{UAV123} benchmarks, giving absolute $AUC$ gains no less than $1.0\%$. Our degraded version simpleRT-MDNet, however, exhibits a little inferior performance to RT-MDNet due to the reason that the update is based solely upon the recent short-term memory. It is noteworthy that our improvement of precision score on \textit{UAV123} is much less than on \textit{OTB-2015}, which can be attributed to the reason that the size of the ground-truth bounding boxes for most of the objects in \textit{UAV123} is much too small so that the tracking deviation from object area does not punish the precision score at 20 pixels too much. This is also why the metric of normalized precision is proposed in~\cite{Muller2018TrackingNet}.

\subsubsection{Comparison with others}
We perform the state-of-the-art comparison on \textit{UAV123} and \textit{VOT2016/VOT2017} by comparing our proposed RLS-RTMDNet to the state-of-the-art CPU-based trackers, \ie, MEEM~\cite{Zhang2014MEEM}, SRDCF~\cite{Danelljan2015SRDCF}, Staple~\cite{Bertinetto2016Staple}, CSRDCF with its real-time version CSRDCF++~\cite{Lukezic2018CSRDCF}, fDSST~\cite{Danelljan2017DSST}, BACF~\cite{Galoogahi2017BACF}, ARCF~\cite{YLi2019ARCF}, AutoTrack~\cite{YLi2020AutoTrack}, ECOhc~\cite{Danelljan2017ECO}, and some deep trackers trained without additionally relying on the \textit{MS-COCO} dataset or recently proposed large-scale tracking datasets, \ie, SiamFC~\cite{Bertinetto2016SiameseFC}, C-COT~\cite{Danelljan2016CCOT}, RT-MDNet~\cite{Jung2018RTMDNet}, MetaSDNet~\cite{Park2018MetaSDNet}, DeepSTRCF~\cite{Li2018STRCF}, UDT with its improved version UDT+~\cite{Wang2019UDT}, TADT~\cite{Li2019TADT}, ECO~\cite{Danelljan2017ECO}, LADCF-R50~\cite{Xu2019LADCF} with a ResNet50 backbone. Results for our degraded version simpleRT-MDNet are also included.
\begin{table}
\small
\setlength{\tabcolsep}{3.2pt} \small
    \fontsize{8pt}{10pt}\selectfont
\centering
\caption{Comparison with Some Top-Performing Deep Trackers on UAV123 in Terms of Mean $OP$ Given Thresholds $0.50$ ($OP_{0.50}$) and $0.75$ ($OP_{0.75}$).}
\begin{tabular}{c!{\vrule width1.2pt}ccccc}
    \Xhline{1.1pt}
    \textit{UAV123} & RT-MDNet & DeepSTRCF & \textbf{RLS-RTMDNet} & TADT & ECO \\
    \Xhline{0.6pt}
    $SR_{0.5}$ & 0.627 & 0.612 & 0.648 & 0.634 & 0.640 \\
    $SR_{0.75}$ & 0.308 & 0.342 & 0.326 & 0.307 & 0.328 \\
    \Xhline{1.1pt}
    \end{tabular}
\label{table:uav1}
\end{table}
\begin{figure}[!t]
\captionsetup{font={small}}
\centering
\subfloat{\includegraphics[width=0.9\linewidth]{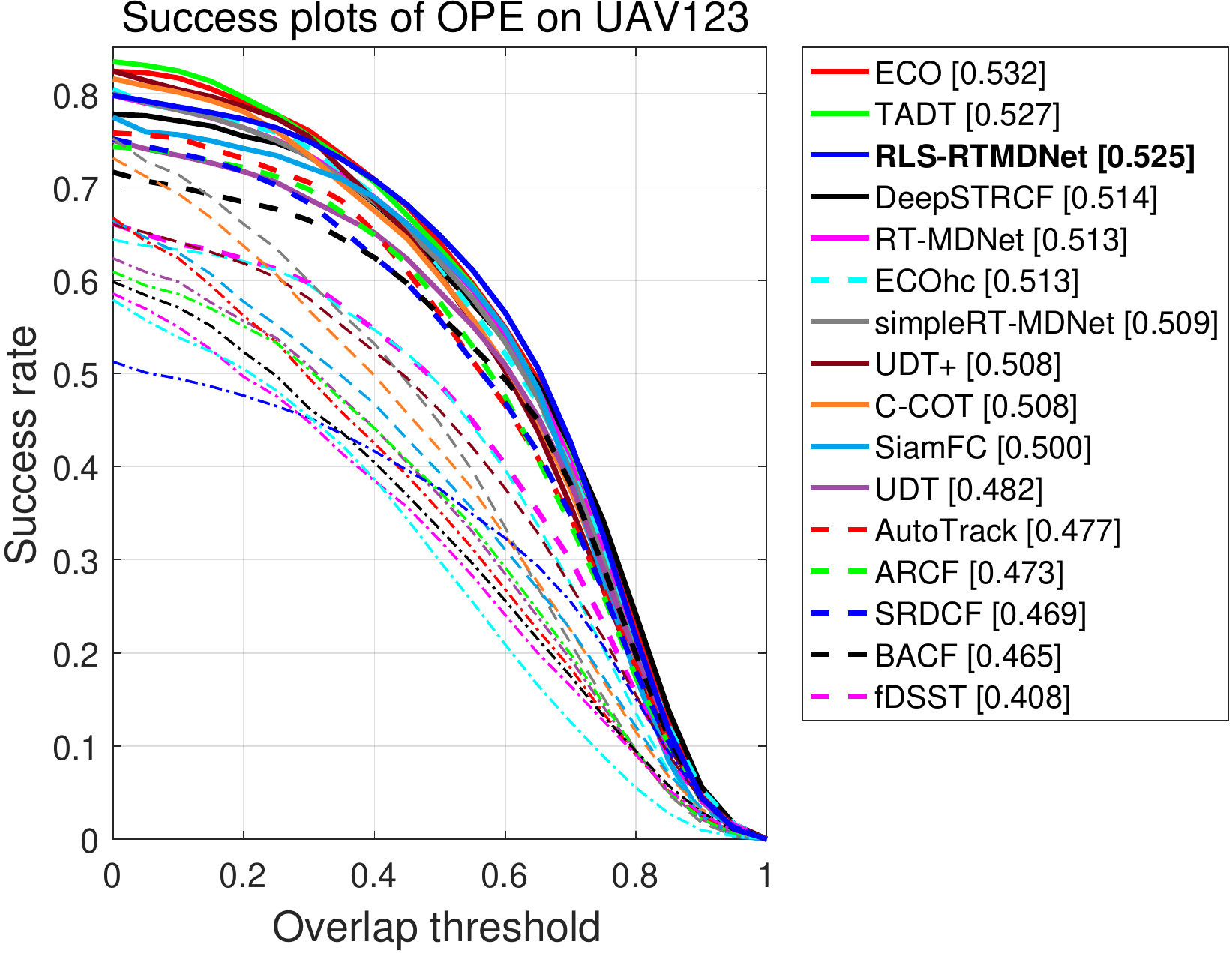}
\label{UAV123}}
\vspace{-2.5mm}
\caption{Success plots for the whole \textit{UAV123} benchmark. We report the average over 50 runs for our tracker RLS-RTMDNet and its corresponding baselines. The legends show the $AUC$ scores. Best viewed in color.} \label{fig:stateofartstudyUAV1}
\vspace{-2.5mm}
\end{figure}

\begin{table*}
	\captionsetup{font={small}}
    \setlength{\tabcolsep}{3pt} \scriptsize
    \fontsize{7pt}{9pt}\selectfont
    \centering
     \caption{Comparison of Our RLS-RTMDNet with Its Corresponding Baselines and Some Related Competing Trackers on the \textit{VOT2016/2017} Benchmarks; the Results Are Reported as $EAO$, $A$, $R_{S}$ ($S=100$) and Real-Time $EAO$ ($rtEAO$). For All These Metrics Except $rtEAO$, the Stochastic Trackers Are Run 15 Times on Each Sequence to Reduce the Variance of Their Results. Best Viewed in Color.}
     \vspace{-2mm}
    \begin{tabular}{cc!{\vrule width1.2pt}ccccccccccccc}
    \Xhline{1.1pt}
    \multicolumn{2}{c!{\vrule width1.2pt}}{\textit{VOT}} & Staple & SiamFC & MEEM & RT-MDNet & simpleRT-MDNet & MetaSDNet & CSRDCF++ & ECOhc & \textbf{RLS-RTMDNet} & CSRDCF & C-COT & ECO & LADCF-R50\\
    \Xhline{1.1pt}
    \multirow{3}{*}{2016} & $EAO\uparrow$ & 0.295 & 0.277 & - & 0.302 & 0.304 & 0.314 & - & 0.322 &  \textcolor{green}{0.336} &  \textcolor{blue}{0.338} & 0.331 &  \textcolor{red}{0.374} & - \\
    & $A\uparrow$ & 0.547 & 0.550 & - & \textcolor{blue}{0.559} & 0.550 & 0.539 & - & 0.542 & \textcolor{red}{0.573} & 0.524 & 0.541 & \textcolor{green}{0.555} & - \\
    & $-\ln R_S\downarrow$ & 0.378 & 0.382 & - & 0.305 & 0.294 & 0.261 & - & 0.303 & \textcolor{green}{0.255} & \textcolor{blue}{0.238} & \textcolor{blue}{0.238} & \textcolor{red}{0.200} & - \\
    \Xhline{1.1pt}
    \multirow{4}{*}{2017} & $EAO\uparrow$ & 0.169 & 0.188 & 0.192 & 0.223 & 0.218 & - & 0.229 & 0.238 & 0.241 & 0.256 & \textcolor{green}{0.267} & \textcolor{blue}{0.280} & \textcolor{red}{0.389} \\
    & $A\uparrow$ & \textcolor{red}{0.530} & 0.503 & 0.463 & \textcolor{green}{0.514} & 0.508 & - & 0.453 & 0.496 & \textcolor{blue}{0.516} & 0.491 & 0.494 & 0.484 & 0.503 \\
    & $-\ln R_S\downarrow$ & 0.688 & 0.585 & 0.534 & 0.450 & 0.464 & - & 0.370 & 0.435 & 0.392 & 0.356 & \textcolor{green}{0.318} & \textcolor{blue}{0.276} & \textcolor{red}{0.159} \\
    & $rtEAO\uparrow$ & 0.170 & \textcolor{blue}{0.182} & 0.072 & 0.162 & 0.154 & - & \textcolor{red}{0.212} & \textcolor{green}{0.177} & 0.150 & 0.099 & 0.058 & 0.078 & 0.066 \\
    \Xhline{1.1pt}
    \end{tabular}
    \label{table:vot1}
    \vspace{-4.5mm}
\end{table*}

\textit{VOT} creates a platform for organizing tracking challenges every year since 2013 by establishing datasets, evaluation measures and toolkits. \textit{VOT2017} departs from \textit{VOT2016} in two aspects: 10 least challenging sequences in \textit{VOT2016} is replaced with new ones; and a new experiment for evaluating real-time performance is introduced in \textit{VOT2017}. Different from \textit{UAV123}, all these \textit{VOT} challenges apply a reset-based methodology in the toolkit. Whenever a failure is detected, the tracker is re-initialized five frames after the failure. Thus, two weakly correlated performance measures can be used: the accuracy ($A$) measures the average overlap between the predicted bounding box and the ground truth computed over the successfully tracked frames; the robustness is estimated by considering the reliability ($R_S$) which shows the probability that the tracker will still successfully track the object up to $S$ frames since the last failure and is computed using the failure rate measure ($R_{\mathrm{fr}}$, the average number of failures) as follows $$R_S = \exp\left(-S\frac{R_{\mathrm{fr}}}{N_{\mathrm{frames}}}\right),$$ where $N_{\mathrm{frames}}$ is the average length of the sequences. 
A more principled expected average overlap ($EAO$) measure is also proposed to measure the expected no-reset average overlap ($AO$) of a tracker run on a short-term sequence, although it is computed from the reset-based methodology.  

We firstly show the comparison on \textit{UAV123} in Figure~\ref{fig:stateofartstudyUAV1} and Table~\ref{table:uav1}. All the entries in this comparison can be roughly categorized into DCF-based, Siamese-style and MDNet-style approaches. In specific, the DCF-based trackers ECO~\cite{Danelljan2017ECO}, DeepSTRCF~\cite{Li2018STRCF} and C-COT~\cite{Danelljan2016CCOT} all use pre-trained deep models to extract powerful features for more robust and accurate tracking. SiamFC~\cite{Bertinetto2016SiameseFC} is the first proposed Siamese-style tracker which can track object very efficiently with online learning abandoned. This is achieved by exploiting end-to-end training along with the advent of deep learning for training tracking models. TADT~\cite{Li2019TADT} further improves the Siamese-style tracking framework by introducing a target-aware feature module to guide the generation of target-active and scale-sensitive features. UDT~\cite{Wang2019UDT} is proposed to explore a feasible unsupervised training approach for deep tracking. AutoTrack~\cite{YLi2020AutoTrack} and ARCF~\cite{YLi2019ARCF} are both DCF-based trackers and dedicated to the real-world low-cost UAV tracking scenario for more efficient tracking, which has been validated in a real-world UAV localization system. As can be seen, the classical DCF-based tracker ECO with deep features and the Siamese-style tracker TADT can both surpass RT-MDNet and simpleRT-MDNet by large margins, while our approach can unveil the power of RT-MDNet to achieve comparable performance with both of them and even the best $SR_{0.5}$ score.

The comparison on \textit{VOT2016/2017} is shown in Table~\ref{table:vot1}.  As similar to the analyses in ablation study, our RLS-RTMDNet always consistently improves upon its baselines (\ie, RT-MDNet and simpleRT-MDNet) by achieving significant gains on \textit{VOT2016/2017}, and our degraded version simpleRT-MDNet achieves similar or a little worse performance than RT-MDNet due to the lack of long-term memory. What's more, our performance gains are larger than MetaSDNet~\cite{Park2018MetaSDNet}, which uses sophisticated meta-training techniques to train the MDNet tracker using more datasets than RT-MDNet.

\subsection{Experimental Analyses of RLS-DiMP}
\label{Section:experimentRLS-DiMP}

We evaluate the proposed RLS estimator-aided online RLS-DiMP tracker thoroughly over the large-scale short-term \textit{TrackingNet-test}~\cite{Muller2018TrackingNet}, \textit{GOT10k-test}~\cite{Huang2019GOT} and \textit{VOT2018/2019}~\cite{Kristan2018VOTconf, Kristan2019VOTconf} tracking benchmarks, and long-term \textit{LaSOT-test}~\cite{Fan2019LaSOT}, \textit{OxUvA-dev}~\cite{Valmadre2018OxUvA} and \textit{TLP}~\cite{Moudgil2019TLP} benchmarks by following rigorously the evaluation protocols. As the off-the-shelf DiMP tracking model is fully trained with the training splits of the \textit{TrackingNet}, \textit{LaSOT}, \textit{GOT10k} and \textit{MS-COCO}~\cite{Lin2014COCO} datasets, we believe this evaluation on more challenging and larger benchmarks can clearly and statistically show the effectiveness of our online learning approach by unveiling the power of the state-of-the-art DiMP tracker.

\subsubsection{Ablation study}
\label{sec:Ablation_RLS-DiMP}

\begin{figure}[!t]
\captionsetup{font={small}}
\centering
\subfloat{\includegraphics[width=0.90\linewidth]{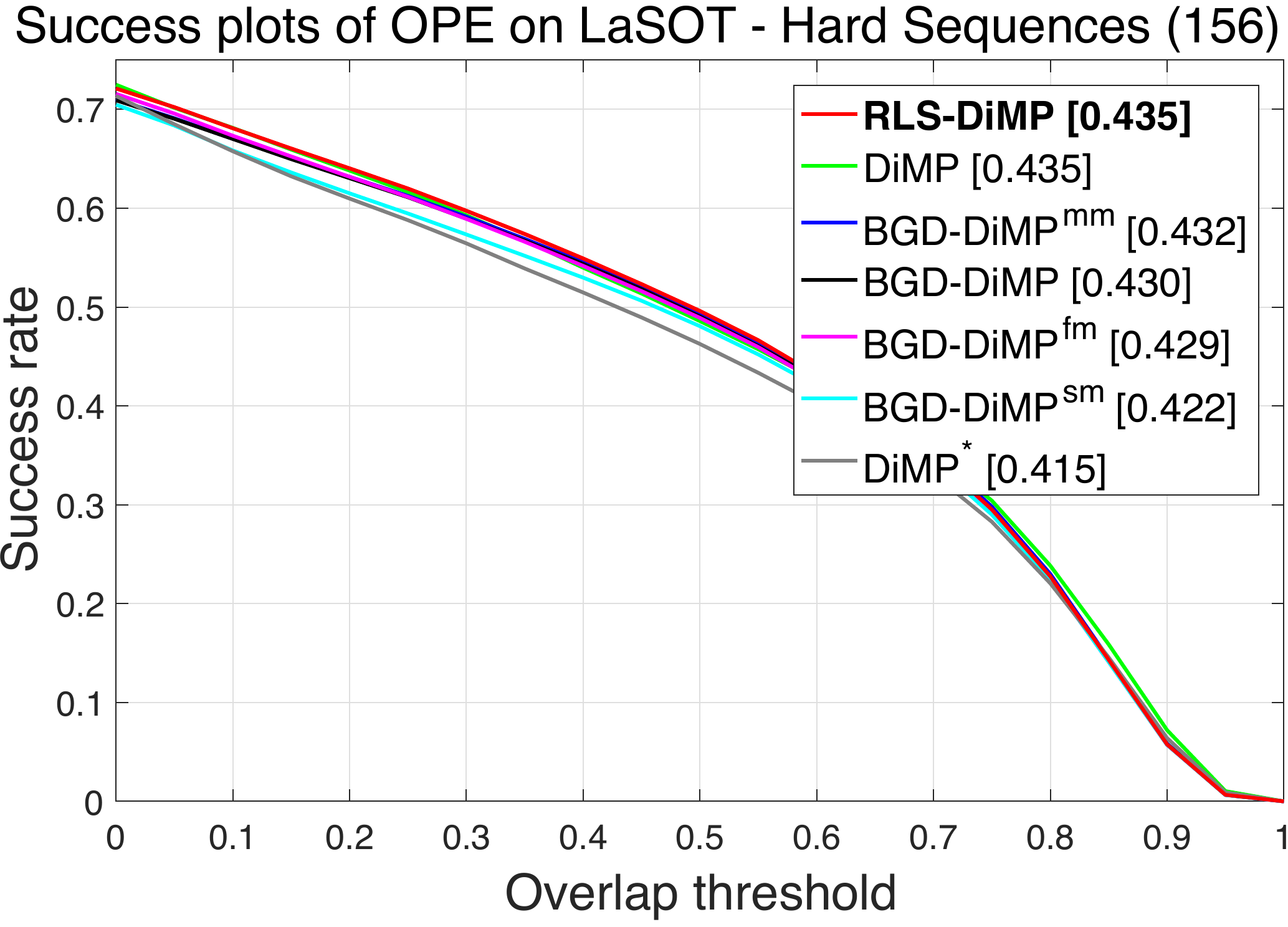}
\label{LaSOT}}
\vspace{-2.5mm}
\caption{Success plots for the hard sequences of the \textit{LaSOT-test} dataset. We report the average over 20 runs for our RLS-DiMP and the re-implemented baselines, except that the publicly available 5-run results are used for DiMP. The legends show the $AUC$ scores.  Best viewed in color.} \label{fig:ablationstudyLaSOT}
\vspace{-2.5mm}
\end{figure}

We start by summarizing the ablation study comparison of our proposed RLS-DiMP with its corresponding baselines on the long-term \textit{LaSOT-test} benchmark in Table~\ref{table:ablationstudy_RLS_DiMP} and Figure~\ref{fig:ablationstudyLaSOT}. \textit{LaSOT} has much longer video sequences with average duration of 84 seconds (2500 frames in average) and each sequence of it comprises at least 1000 frames, which is dedicated to the long-term tracking principle. It is split into training and testing subsets, and the testing subset consists of 280 sequences with $690K$ frames. It also uses the same evaluation protocol to \textit{OTB-2015}. However, due to the fact that the precision metric is sensitive to the resolution of the images and the size of the bounding boxes, a metric of normalized precision~\cite{Muller2018TrackingNet} over the size of the ground-truth bounding box can be exploited and the trackers can be then ranked using the $AUC$ for normalized precision between 0 and 0.5.

As claimed in Sec.~\ref{Section:setup}, we add the baselines using the EMA strategy for saving historical information in addition to BGD-DiMP and DiMP$^*$. Due to the reason that our memory retention is applied to improve the vanilla BGD in BGD-DiMP and BGD-DiMP performs better than the original DiMP on \textit{LaSOT-test}, we hence apply the EMA strategy on the baseline BGD-DiMP to construct new baselines. In specific, let $\alpha$ denote the EMA coefficient, and then the approximate optimum value $\widehat{\mathbf{W}}_{n}^{appr}$ after the vanilla BGD-based optimization at each update stage of BGD-DiMP can be further updated as follows
$$
\widehat{\mathbf{W}}_{n}^{appr} \longleftarrow \left(1-\alpha\right)\widehat{\mathbf{W}}_{n-1}^{appr} + \alpha\widehat{\mathbf{W}}_{n}^{appr}~,
$$
where $\alpha \in [0, 1]$ and represents the degree of weighting decrease. A higher $\alpha$ discounts older parameters faster.

The commonly used slow-moving average strategy as mentioned in Sec.~\ref{Section:setup} always sets $\alpha$ to be less than 0.05 so that the online tracking model's parameters or the network representation in self-supervised learning evolve smoothly. However, applying this kind of EMA strategy on BGD-DiMP makes it prone to lower adaptive capacity while saving historical information, leading to decrease of performance on \textit{LaSOT-test}. It is also impractical to do an exhaustive grid search to obtain the optimum value of $\alpha$. So we roughly set three baselines as follows: BGD-DiMP$^{\rm{sm}}$ for discounting older parameters in a slow-moving way with $\alpha$ set to 0.01, BGD-DiMP$^{\rm{mm}}$ in a moderate-moving way with $\alpha$ set to 0.5, and BGD-DiMP$^{\rm{fm}}$ in a fast-moving way with $\alpha$ set to 0.99. This practice can also be observed in LADCF~\cite{Xu2019LADCF} that the EMA coefficient is set to 0.95 and 0.13 for hand-crafted and deep LADCF respectively. It is noteworthy that the baseline BGD-DiMP can be seen as a special case of applying the EMA strategy by setting $\alpha$ to 1.0.

\begin{table}
\small
\setlength{\tabcolsep}{2.6pt} \small
    \fontsize{8pt}{10pt}\selectfont
\centering
\caption{Ablation Study Comparison of RLS-DiMP with Its Corresponding Baselines on the Long-Term \textit{LaSOT-test} Benchmark. $Prec.$, $Norm.~Prec.$ and $Succ.$ Denote Precision Score at 20 Pixels, Normalized Precision Score and $AUC$ of Success Plot Respectively. The Average Over 20 Runs Is Reported for These DiMP-Style Trackers Except That We Report the Publicly Available 5-Run Results (Indicated by $^{\dagger}$) for DiMP.}
\begin{tabular}{c!{\vrule}ccc!{\vrule}c!{\vrule}c}
\toprule
\multirow{2}{*}{Trackers} & \multicolumn{3}{c!{\vrule}}{\textit{LaSOT-test}} & Running & GPU Memory\\
\cmidrule(lr){2-4}
&  $Prec.$ &  $Norm.~Prec.$ &  $Succ.$  & Speed & Consumption \\
\midrule
DiMP & 0.567$^{\dagger}$  & 0.650$^{\dagger}$ & 0.569$^{\dagger}$ & $\sim$41 \emph{fps} & \multirow{2}{*}{$\sim$2211 MiB} \\
DiMP$^*$ & 0.556  & 0.636 & 0.557 & $\sim$40.6 \emph{fps} &  \\
\midrule
BGD-DiMP & 0.575 & 0.673 & 0.574 & \multirow{4}{*}{$\sim$41 \emph{fps}}& \multirow{4}{*}{$\sim$2211 MiB}  \\
BGD-DiMP$^{\rm{fm}}$ & 0.574  & 0.673 & 0.574 &  & \\
BGD-DiMP$^{\rm{mm}}$ & 0.576  & 0.675  & 0.575 &  &  \\
BGD-DiMP$^{\rm{sm}}$ & 0.558  & 0.659  & 0.561 &  & \\
\midrule
\textbf{RLS-DiMP} & 0.578 & 0.676  & 0.577 & $\sim$40 \emph{fps} & $\sim$2339 MiB  \\
\end{tabular}
\label{table:ablationstudy_RLS_DiMP}
\end{table}

It is clearly and statistically shown in Table~\ref{table:ablationstudy_RLS_DiMP} that our online learning approach can also unveil the power of the state-of-the-art DiMP tracker without incurring too much extra computational and memory cost. As \textit{LaSOT-test} is dedicated to the long-term tracking principle, increasing the SD optimizer recursion number from two to five in DiMP$^*$ may aggravate the overfitting issue caused by fast converging to a \textit{local} point with respect to the fixed-sized sample set, and harm the overall performance on \textit{LaSOT-test}. By experimentally setting an appropriate small learning rate for BGD-DiMP, we can reduce overfitting and achieve superior results compared to the original DiMP tracker. Thanks to the retention of historical information by exploiting the EMA strategy, the baseline BGD-DiMP$^{\rm{mm}}$ with the coefficient $\alpha$ roughly set to 0.5 can further enhance this superiority. We believe we can use some art to choose an optimum value of $\alpha$ for this EMA strategy to make a comprise between adapting quickly and maintaining robustness, and thus improve BGD-DiMP by a larger margin. However, our proposed online learning approach has been demonstrated theoretically to be able to retain the memory in a more elegant way, and thus reduce the risk of overfitting as well as maintain adaptive capacity. That means our RLS-DiMP can improve the performance on \textit{LaSOT-test} without requiring fine-grained searching for hyper-parameters by hand.

Since some sequences in \textit{LaSOT-test} are less challenging for the DiMP-style trackers to achieve good performance, we selected some difficult ones for an in-depth analysis as shown in Figure~\ref{fig:ablationstudyLaSOT}, which also makes the abovementioned phenomena more clearly understood. As each sequence was evaluated with several runs by each entry in Table~\ref{table:ablationstudy_RLS_DiMP}, we can define the sequence for which the $AUC$ scores of different runs by our RLS-DiMP are all above $0.75$ as easy sequence and remove it from our evaluation to achieve the success plots on the remaining hard sequences of \textit{LaSOT-test}, leading to 156 hard sequences in total. From Figure~\ref{fig:ablationstudyLaSOT} we can see that, our RLS-DiMP improves the baseline BGD-DiMP with an absolute $AUC$ gain of $0.5\%$ on these hard sequences, which is computed based on 20 runs of the test sequences due to the stochastic nature of DiMP-style trackers. By comparing the performance gains between Table~\ref{table:ablationstudy_RLS_DiMP} and Figure~\ref{fig:ablationstudyLaSOT}, we can also find that introducing memory retention mainly helps to improve the tracking performance on the more challenging sequences, which is preferable as we aim to handle the challenging factors in tracking. It is noteworthy that only concentrating on saving historical information without maintaining adaptive capacity also harm the overall performance on \textit{LaSOT-test}.

\subsubsection{Comparison with others}
We perform the state-of-the-art comparison on \textit{TrackingNet-test}, \textit{GOT10k-test}, \textit{LaSOT-test}, \textit{OxUvA-dev}, \textit{TLP} and \textit{VOT2018/2019} by comparing our proposed RLS-DiMP to some short-term trackers, \ie, SiamFC~\cite{Bertinetto2016SiameseFC}, MDNet~\cite{Nam2016MDNet}, ECO~\cite{Danelljan2017ECO}, DaSiamRPN~\cite{Zhu2018DaSiamRPN}, ATOM~\cite{Danelljan2019ATOM}, GradNet~\cite{Li2019GradNet}, SiamMask~\cite{Wang2019SiamMask}, UpdateNet-DaSiamRPN~\cite{Zhang2019UpdateNet}, SiamRPN++~\cite{Li2019SiamRPN++}, SiamFC++-GoogLeNet~\cite{Xu2020SiamFC++}, D3S~\cite{Lukezic2020D3S}, FCOS-MAML~\cite{Wang2020Retina-MAML}, SiamBAN~\cite{Chen2020SiamBAN}, ROAM++~\cite{Yang2020ROAM}, PrDiMP~\cite{Danelljan2020PrDiMP}, and long-term trackers, \ie, SPLT~\cite{Yan2019SPLT}, GlobalTrack~\cite{Huang2020GlobalTrack}, Siam R-CNN~\cite{Voigtlaender2020SiamRCNN}, LTMU~\cite{Dai2020LTMU}. Most of these competitors are trained additionally relying on the \textit{MS-COCO} dataset or recently proposed large-scale tracking datasets. Results for the re-implemented baselines are also included.
\begin{table*}[!t]
	
	\begin{center}
		\fontsize{8pt}{10pt}\selectfont
		\caption{Comparison of Our RLS-DiMP with Its Corresponding Baselines and Some State-of-the-Art Competing Trackers on \textit{TrackingNet-test}, \textit{GOT10k-test}, \textit{LaSOT-test}, \textit{OxUvA-dev}, \textit{TLP}; the Results of Our RLS-DiMP and Its Corresponding Baselines Are Reported Based on 20 Runs on Each Sequence. $Prec.$ and $Succ.$ Denote Precision Score at 20 Pixels and $AUC$ of Success Plot Respectively. ($\cdot$, $\cdot$) Denotes the Number of Video Sequences and Frames for Evaluation in the Corresponding Dataset.}
		\begin{threeparttable}
		\begin{tabular}{llccccccccccc}
			
			\toprule
			\multirow{3}{*}{Trackers}&\multirow{3}{*}{Year}&\multicolumn{2}{c}{\textit{TrackingNet-test}}&\multicolumn{3}{c}{\textit{GOT10k-test}}&\multicolumn{2}{c}{\textit{LaSOT-test}}&\multicolumn{2}{c}{\textit{OxUvA-dev}}&\multicolumn{2}{c}{\textit{TLP}}\cr
			& & \multicolumn{2}{c}{($511, 226K$)}&\multicolumn{3}{c}{($420, 56K$)}&\multicolumn{2}{c}{($280, 685K$)}&\multicolumn{2}{c}{($200, 547K$)}&\multicolumn{2}{c}{($50, 676K$)}\cr
			\cmidrule(lr){3-4} \cmidrule(lr){5-7} \cmidrule(lr){8-9} \cmidrule(lr){10-11} \cmidrule(lr){12-13}
			& &$Succ.$ &$Prec.$&$AO$&$SR_{0.5}$&$SR_{0.75}$&$Succ.$ &$Prec.$&$TPR$&$TNR$&$Succ.$ &$Prec.$\cr
			\cmidrule(lr){1-13}
			\multicolumn{2}{l}{\textit{Short-term Trackers}}&&&&&&&&&&&\vspace*{5pt}\cr
			SiamFC&2016&0.571&0.663&0.348&0.353&0.098&0.336&0.339&0.391&0.0&0.237&0.280 \cr
			MDNet&2016&0.606&0.565&0.299&0.303&0.099&0.397&0.373&0.472&0.0&0.372&0.381  \cr
			ECO&2017&0.554&0.492&0.316&0.309&0.111&0.324&0.301&-&-&0.205&0.211 \cr
			ATOM&2019&0.703&0.648&0.556&0.634&0.402&0.515&0.505&-&-&-&-\cr
			UpdateNet-DaSiamRPN&2019&0.677&0.625&-&-&-&0.475&-&-&-&-&-\cr
			SiamRPN++&2019&0.733&0.694&0.517&0.616&0.325&0.496&0.491&-&-&-&-\cr
			SiamFC++-GoogLeNet&2020&0.754&0.705&0.595&0.695&0.479&0.544&0.547&-&-&-&- \cr
			D3S&2020&0.728&0.664&0.597&0.676&0.462&-&-&-&-&-&- \cr
			FCOS-MAML&2020&0.757&0.725&-&-&-&0.523&0.531&-&-&-&-\cr
			SiamBAN&2020&-&-&-&-&-&0.514&0.521&-&-&-&-\cr
			ROAM++&2020&0.670&0.623&0.465&0.532&0.236&0.447&0.445&-&-&-&-\cr
			PrDiMP&2020&0.758&0.704&0.634&0.738&0.543&0.598&0.608&-&-&-&-\cr
			\cmidrule(lr){1-13}
			\multicolumn{2}{l}{\textit{Long-term Trackers}}&&&&&&&&&&&\vspace*{5pt}\cr
			SPLT&2019&-&-&-&-&-&0.426&0.396&0.498&0.776&0.416&0.403\cr
		   GlobalTrack&2020&0.704&0.656&-&-&-&0.521&0.527&0.574&0.633&0.520&0.556\cr
			Siam R-CNN&2020&0.812&0.800&0.649&0.728&0.597&0.648&-&0.701&0.745&-&-\cr
			LTMU&2020&-&-&-&-&-&0.572&0.572&0.749&0.754&0.571&0.608\cr
			\cmidrule(lr){1-13}
			\multicolumn{2}{l}{\textit{Our Implementations}}&&&&&&&&&&&\vspace*{5pt}\cr
			DiMP&2019&0.740&0.689&0.609&0.709&0.491&0.569$^{\dagger}$&0.567$^{\dagger}$&0.727&0.0&0.502&0.513 \cr
			DiMP$^*$&2019&0.742&0.690&0.608&0.709&0.490&0.557&0.556&0.714&0.0&0.497&0.507 \cr
			BGD-DiMP&Ours&0.741&0.689&0.606&0.713&0.495&0.574&0.575&0.731&0.0&0.508&0.506 \cr
			\textbf{RLS-DiMP}&Ours&0.740&0.688&0.611&0.713&0.492&0.577&0.578&0.735&0.0&0.519&0.518 \cr
			\bottomrule
		\end{tabular}
		\begin{tablenotes}
        \footnotesize
        \item[$\dagger$] We report the publicly available 5-run results for DiMP.
      \end{tablenotes}
      \end{threeparttable}
		\vspace{-4mm}
		\label{tab:RLS_DiMPresults}
	\end{center}
\end{table*}

As most researchers have designed tracking methods tailored to the short-term scenario, which is poorly representative of practitioners’ needs, \textit{TLP} and \textit{OxUvA} are recently proposed aiming to address this disparity. The former compiles 50 HD videos from real world scenarios, encompassing a duration of over 400 minutes (676K frames), while the latter comprises 366 sequences spanning 14 hours of video. \textit{OxUvA} is also split into \textit{dev} and \textit{test} sets of 200 and 166 tracks respectively, where the trackers can be evaluated through an online server. What's more, \textit{OxUvA} contains an average of 2.2 absent labels per track and at least one labelled disappearance in 52$\%$ of the tracks, which enables to evaluate trackers in terms of $TPR$ (fraction of present objects that are reported present and correctly located) and $TNR$ (fraction of absent objects that are reported absent). A problem with \textit{OxUvA} is that it does not provide dense annotations in consecutive frames, \ie, each video in \textit{OxUvA} is annotated every 30 frames, ignoring rich temporal context between consecutive frames for developing a tracker. So the aforementioned long-term \textit{LaSOT} benchmark with high-quality dense annotations is proposed to bridge this gap, leading to the largest high-quality dense tracking benchmark which also provides a training subset with 2.8 million boxes for developing a tracker.

Before \textit{LaSOT}, \textit{TrackingNet} is the first large-scale tracking benchmark by simultaneously providing the training subset for developing a tracker and the test subset for evaluating different trackers. An online server is also provided for this evaluation. However, \textit{TrackingNet} is manually labeled at 1 \textit{fps} while its all other annotations are automatically generated using correlation filter-based tracking. The most recently proposed \textit{GOT10k} benchmark offers a much wider coverage of object classes and first introduces the one-shot protocol to avoid evaluation bias towards seen familiar object classes, \ie, the training and test classes are zero-overlapped, which can be used to evaluate the generalization of different trackers. This is very different from \textit{LaSOT} and \textit{TrackingNet}. Besides $SR$, \textit{GOT10k} also exploits $AO$ to denote the average of overlaps between all ground-truth and estimated bounding boxes, which is proved to be equivalent to the $AUC$ metric.

We firstly show the comparison on these large-scale tracking benchmarks in Table~\ref{tab:RLS_DiMPresults}. Since \textit{GOT10k} requires the trackers to be trained on its training set with no extra training data used and the DiMP tracking model following this protocol is not public available, we thus re-trained the DiMP tracking model for the evaluation on \textit{GOT10k}. Our implementation results of RLS-DiMP and its corresponding baselines are reported based on 20 runs on each sequence.  It clearly shows that our online learning approach can effectively unveil the power of DiMP and surpass the baseline BGD-DiMP by achieving better performance on all the long-term \textit{LaSOT-test}, \textit{OxUvA-dev} and \textit{TLP} datasets. This is desirable as our framework is designed for memory retention and overfitting alleviation. What's more, the original DiMP tracker based on the SD optimizer performs worse on all these long-term datasets, especially with a significant performance decrease for DiMP$^*$. This is consistent with the analysis in Sec.~\ref{sec:Ablation_RLS-DiMP} that DiMP$^*$ may aggravate the overfitting issue and harm the overall performance in long-term tracking. 

However, short-term tracking may prefer overfitting more than our memory retention, which makes DiMP$^*$ perform best on \textit{TrackingNet-test} in our implementations. This phenomenon can also be attributed to the reason that the object classes between training and test sets in \textit{TrackingNet} are fully overlapped with close distribution, and the DiMP tracking model trained with \textit{TrackingNet-train} used may lead to biased evaluation results towards familiar objects on \textit{TrackingNet-test}. This is further validated by the  evaluation on \textit{GOT10k-test}, which shows that our online learning approach can make the DiMP tracking model generalize to the unseen object classes better than the other SD and vanilla BGD optimization methods in the one-shot tracker evaluation protocol.

Table~\ref{tab:RLS_DiMPresults} also shows that the novel ROAM++~\cite{Yang2020ROAM} and UpdateNet-DaSiamRPN~\cite{Zhang2019UpdateNet} trackers with corresponding LSTM-based and ConvNet model-based meta-learners exhaustively offline trained achieve substantially inferior results compared to our RLS-DiMP because of their exploited modest baselines. Our approach, however, can effectively unveil the power of the top-performing baseline tracker DiMP without incurring too much extra computational cost. In spite of the dominant performance on \textit{TrackingNet-test}, \textit{GOT10k-test} and \textit{LaSOT-test}, Siam R-CNN~\cite{Voigtlaender2020SiamRCNN} still generalizes not well on \textit{OxUvA-dev} compared with LTMU~\cite{Dai2020LTMU} due to the missing of online learning. 

\begin{table*}
	\captionsetup{font={small}}
    \setlength{\tabcolsep}{2.4pt} \scriptsize
    \fontsize{7pt}{9pt}\selectfont
    \centering
     \caption{Comparison of Our RLS-DiMP with Its Corresponding Baselines and Some State-of-the-Art Competing Trackers on \textit{VOT2018/2019}; the Results Are Reported as $EAO$, $A$, $R_{S}$ ($S=100$) and Real-Time $EAO$ ($rtEAO$). For All These Metrics Except $rtEAO$, the Stochastic Trackers Are Run 15 Times on Each Sequence to Reduce the Variance of Their Results. Best Viewed in Color.}
     \vspace{-2mm}
    \begin{tabular}{cc!{\vrule width1.2pt}ccccccccccccc}
    \Xhline{1.1pt}
    \multicolumn{2}{c!{\vrule width1.2pt}}{\multirow{2}{*}{\textit{VOT}}} & GradNet & DaSiamRPN & ROAM++ & SiamMask & UpdateNet & MAML & SiamRPN++ & SiamFC++ & BGD-DiMP & DiMP & DiMP$^*$ & \textbf{RLS-DiMP} & PrDiMP  \\
    & & &  &  &  & (-DaSiamRPN) & (-FCOS) &  & (-GoogLeNet)  &  &  &  &   &   \\
    \Xhline{1.1pt}
    \multirow{4}{*}{2018} & $EAO\uparrow$ & 0.247 & 0.326 & 0.380 & 0.384 & 0.393 & 0.392 & 0.415 & \textcolor{green}{0.426} &  0.415 & 0.423 & 0.425 &  \textcolor{red}{0.443} & \textcolor{blue}{0.442}\\
    & $A\uparrow$ & 0.507 & 0.570 & 0.543 & \textcolor{green}{0.609} & - & \textcolor{red}{0.635} & 0.600 & 0.587 & 0.598 & 0.602 & 0.604 & 0.602 & \textcolor{blue}{0.618}\\
    & $-\ln R_S\downarrow$ & 0.375 & 0.337 & 0.195 & 0.281 & - & 0.220 & 0.234 & 0.183 & \textcolor{blue}{0.163} & 0.170 & \textcolor{green}{0.164} & \textcolor{red}{0.138} & 0.165\\
    & $rtEAO\uparrow$ & - & - & - & 0.384 & - & - & 0.415 & - & 0.393 & \textcolor{green}{0.418} & \textcolor{red}{0.450} & \textcolor{blue}{0.422} & -\\
    \Xhline{1.1pt}
    \multirow{4}{*}{2019} & $EAO\uparrow$ & - & - & 0.281 & 0.287 & - & 0.295 & 0.285 & - & 0.324 & \textcolor{green}{0.357} & \textcolor{blue}{0.360} & \textcolor{red}{0.371} & -\\
    & $A\uparrow$ & - & - & 0.561 & \textcolor{green}{0.594} & - & \textcolor{red}{0.637} & \textcolor{blue}{0.599} & - & \textcolor{blue}{0.599} & 0.588 & 0.583 & \textcolor{green}{0.594} & -\\
    & $-\ln R_S\downarrow$ & - & - & 0.438 & 0.461 & - & 0.421 & 0.482 & - & 0.323 & \textcolor{green}{0.284} & \textcolor{blue}{0.280} & \textcolor{red}{0.265} & -\\
    & $rtEAO\uparrow$ & - & - & 0.110 & 0.287 & - & - & 0.285 & - & 0.328 & \textcolor{blue}{0.353} & \textcolor{green}{0.340} & \textcolor{red}{0.354} & - \\
    \Xhline{1.1pt}
    \end{tabular}
    \vspace{-2.5mm}
    \label{table:vot2}
\end{table*}

We also report the comparison results on \textit{VOT2018/2019} short-term challenges in Table~\ref{table:vot2}. Since the \textit{VOT2017} dataset had not saturated by the time the \textit{VOT2018}'s challenges were held, the dataset was used unchanged in the \textit{VOT2018} short-term challenge. In the \textit{VOT2019} short-term challenge, it had been decided to refresh the public dataset by replacing 12 least difficult sequences with more challenging ones. As can be seen in Table~\ref{table:vot2}, our RLS-DiMP surpasses all its corresponding baselines on both \textit{VOT2018} and \textit{VOT2019} by significantly reducing the number of tracking failures thanks to the proposed memory retention mechanism.

\subsection{Complexity Analyses}
\label{Section:experimentComplexity}

In Table~\ref{table:ablationstudy_RLS_RTMDNet} and Table~\ref{table:ablationstudy_RLS_DiMP}, we also show some detailed comparisons regarding the efficiency of our proposed approach by reporting the running speed and GPU memory consumption for both our proposed methods and the re-implemented baselines. It can be seen that our approach enables us to unveil the power of RT-MDNet and DiMP trackers without incurring too much extra computational cost. 

From Section~\ref{Sec:RLS-Formulation} we can find that there are some matrix multiplication operations in our approach, which leads to the runtime of ${\mathcal {O}}(p^2)$ with respect to the dimension $p$ of the input $\mathbf{x}$. However, all these operations are executed by the GPU along with the other neural network operations so that the final tracking speed is only slightly affected. In order to achieve memory retention, our approach requires the trackers to maintain the matrix $\mathbf{P}_{n}$ in the whole tracking phase and also update it with above matrix multiplication operations, which results in the ${\mathcal {O}}(p^2)$ memory requirement on the GPU. This factor has little impact on the GPU memory consumption for our RLS-RTMDNet while affecting RLS-DiMP too much due to the $img2col$ operation and its subsequent high-dimensional input for online learning ($p = 4\times 4 \times 512$). We thus use the half-float precision for maintaining and updating $\mathbf{P}_{n}$, which reduces the RLS's GPU memory usage significantly.
Finally, we can use some code optimization techniques to further improve the efficiency, though it is outside the scope of this work.

\section{Conclusion}

We present a recursive least-squares estimator-aided network online learning approach that allows in-built memory retention mechanism for improving few-shot online adaptation in tracking. We apply it to two networks in the online learning families for tracking: the MLPs-based as in RT-MDNet and the ConvNets-based as in DiMP. The experiments demonstrate its efficiency in reducing the risk of overfitting through memory retention while maintaining adaptive capacity. This can be attributed to the reason that this recursive learning approach enables the optimization to approximately converge to a {\it global} point with respect to all the historical training samples ever seen including the discarded old data. An interesting direction for future work is to integrate it into the meta-learning-based offline training framework to gain further improvements.

\ifCLASSOPTIONcompsoc
  \section*{Acknowledgments}
\else
   regular IEEE prefers the singular form
  \section*{Acknowledgment}
\fi

\HLredrevision{The authors would like to thank the anonymous reviewers for their valuable comments and suggestions.}




%
\bibliographystyle{IEEEtran}
\bibliography{IEEEabrv,mybib}

%
%

%

\vspace{-15mm}

\begin{IEEEbiography}[{\includegraphics[width=1in,height=1.2in,clip,keepaspectratio]{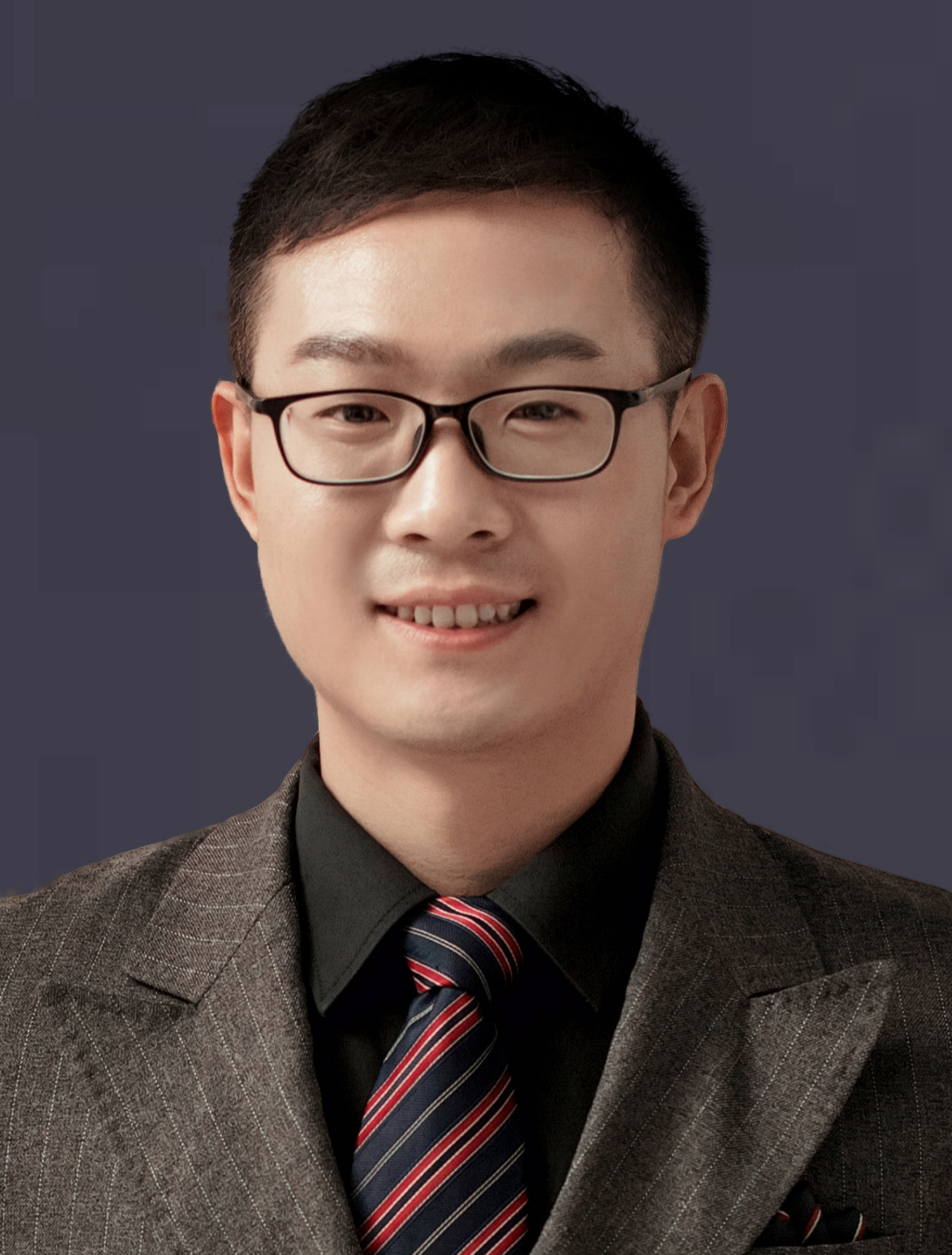}}]{Jin Gao}
	received the BS degree from the Beihang University, Beijing, China, in 2010, and the PhD degree from the University of Chinese Academy of Sciences (UCAS), in 2015. Now he is an associate professor with the National Laboratory of Pattern Recognition (NLPR), Institute of Automation, Chinese Academy of Sciences (CASIA). His research interests include visual tracking, autonomous vehicles, and service robots.
\end{IEEEbiography}
\vspace{-17mm}

\begin{IEEEbiography}[{\includegraphics[width=1in,height=1.2in,clip,keepaspectratio]{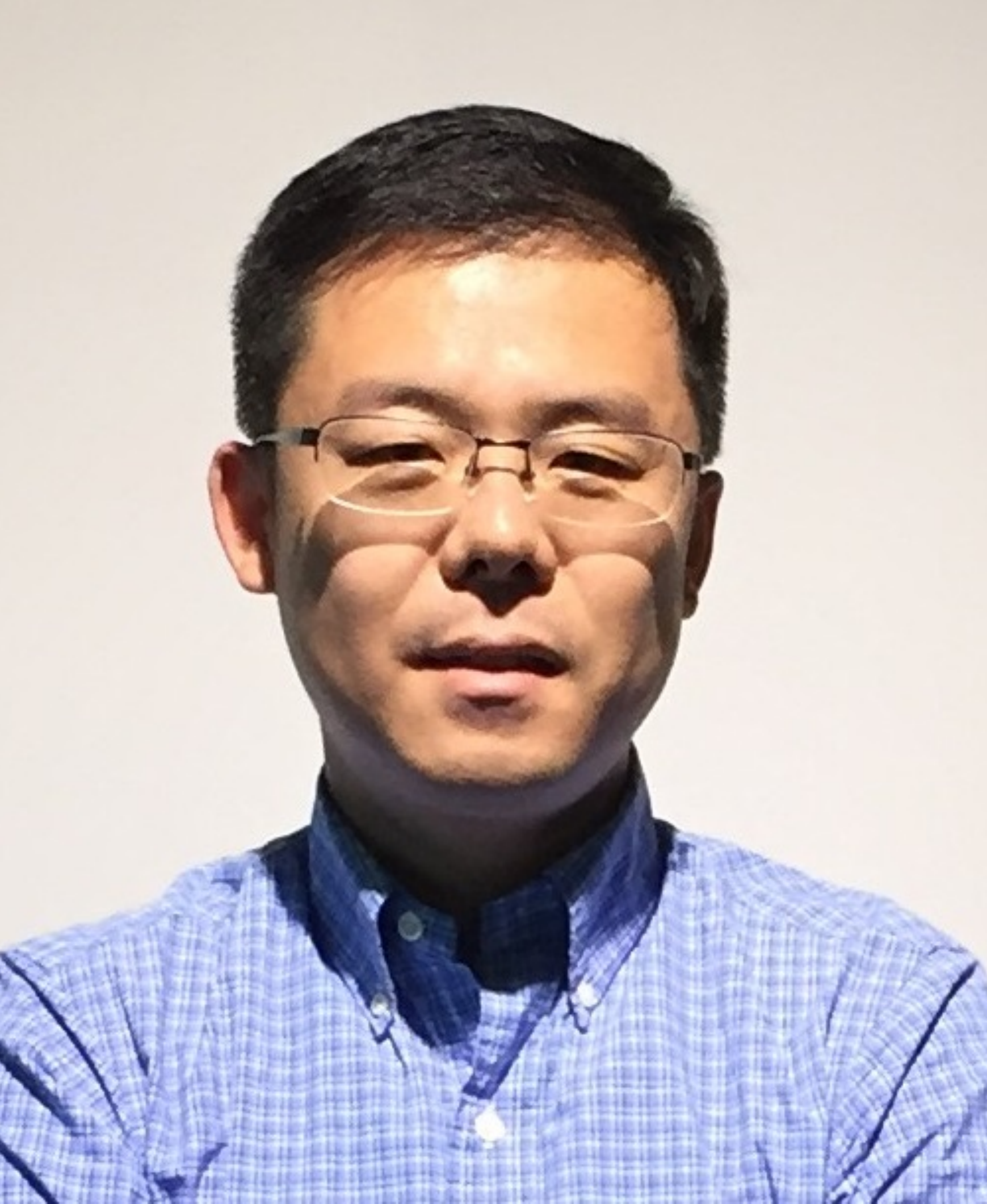}}]{Yan Lu}
	received his PhD degree in computer science from Harbin Institute of Technology, China, in 2004. He joined in Microsoft Research Asia in 2004, where he is now the Partner Research Manager of Media Computing Group, leading the development of core technologies in the fields of real-time communication, computer vision, video analytics, audio enhancement, virtualization, and mobile-cloud computing. From 2001 to 2004, Yan Lu was the team lead of video coding group in the JDL Lab, Institute of Computing Technology, China. From 1999 to 2000, he was with the City University of Hong Kong as a research assistant. His research interests include image and video coding, computer vision, audio and speech, multimedia system, networking, and remote computing. Yan Lu has published 100+ papers and holds 30+ granted US patents in the field of multimedia and computer vision.
\end{IEEEbiography}
\vspace{-17mm}

\begin{IEEEbiography}[{\includegraphics[width=1in,height=1.2in,clip,keepaspectratio]{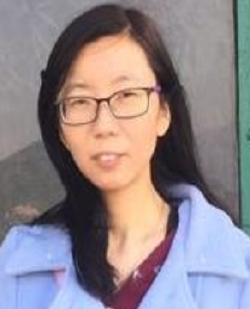}}]{Xiaojuan Qi}
	received her B.Eng degree in Electronic Science and Technology at Shanghai Jiao Tong University (SJTU), in 2014, and the PhD degree in Computer Science and Engineering from the Chinese University of Hong Kong in 2018. She is now an assistant professor in the University of Hong Kong.
\end{IEEEbiography}
\vspace{-17mm}

\begin{IEEEbiography}[{\includegraphics[width=1in,height=1.2in,clip,keepaspectratio]{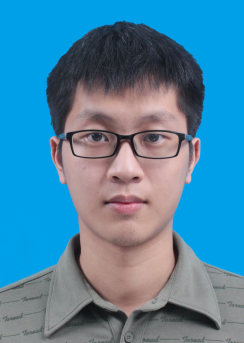}}]{Yutong Kou}
	expects to earn the BS degree at School of Computer Science and Technology, Huazhong University of Science and Technology, Wuhan, China, in June 2021. He will be enrolled in the University of Chinese Academy of Science (UCAS) to pursue the Master's degree at Institute of Automation, Chinese Academy of Sciences (CASIA). His research interests include both theory and applications of single object tracking.
\end{IEEEbiography}
\vspace{-17mm}

\begin{IEEEbiography}[{\includegraphics[width=1in,height=1.2in,clip,keepaspectratio]{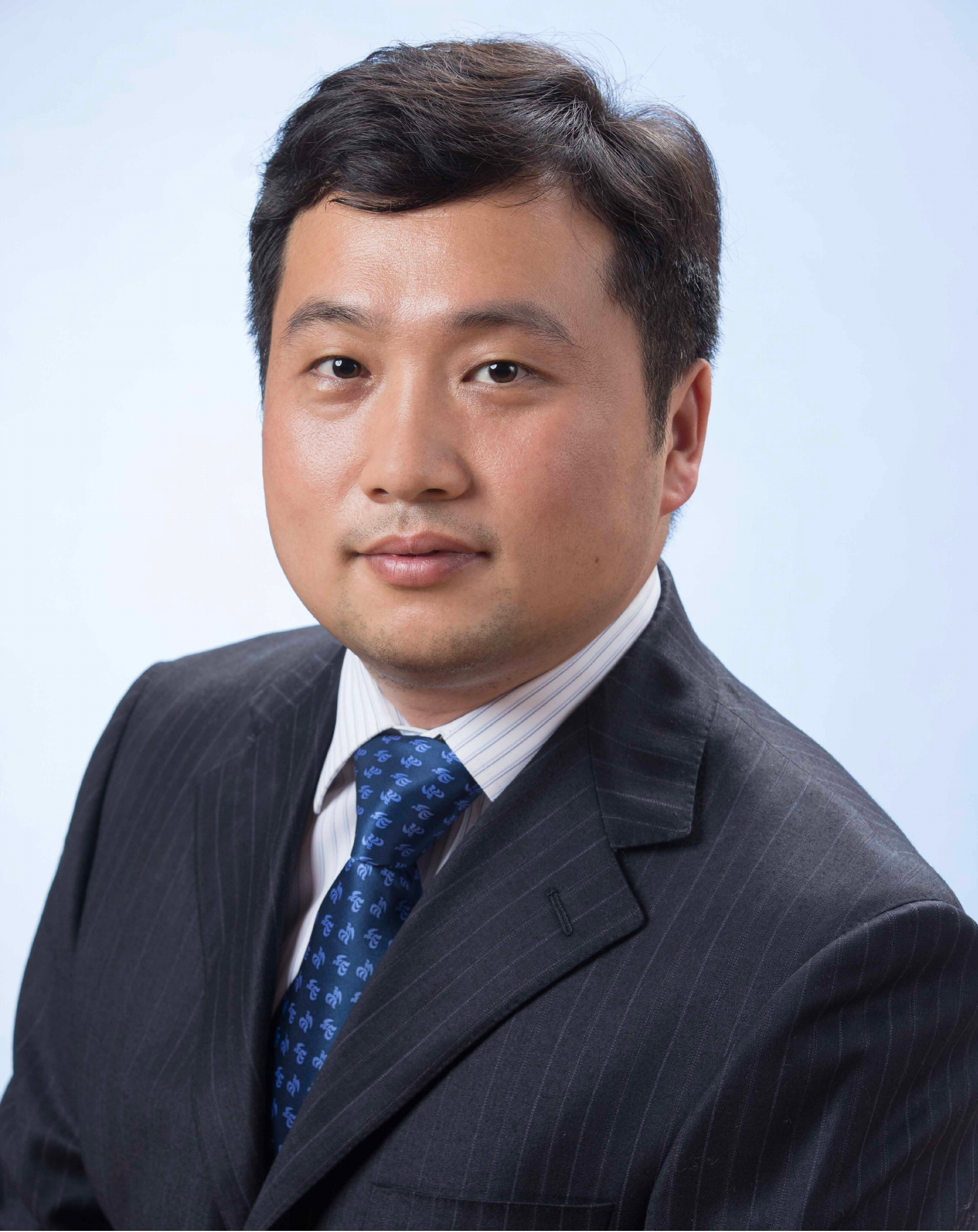}}]{Bing Li}
	received the PhD degree from the Department of Computer Science and Engineering, Beijing Jiaotong University, Beijing, China, in 2009. From 2009 to 2011, he worked as a Postdoctoral Research Fellow with the National Laboratory of Pattern Recognition, Institute of Automation, Chinese Academy of Sciences (CASIA), Beijing. He is currently a Professor with CASIA. His current research interests include computer vision, color constancy, visual saliency detection, multi-instance learning, and data mining.
\end{IEEEbiography}
\vspace{-17mm}

\begin{IEEEbiography}[{\includegraphics[width=1in,height=1.2in,clip,keepaspectratio]{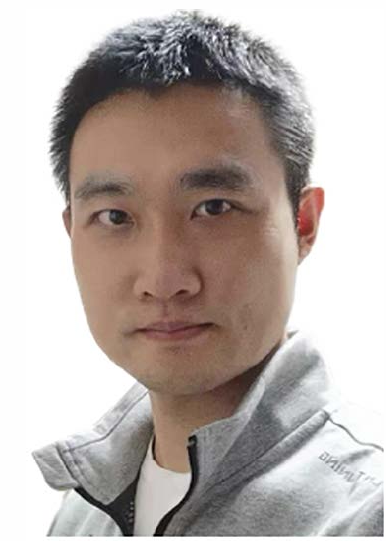}}]{Liang Li}
	received the PhD degree from the Fourth Military Medical University (FMMU), in 2017. Now he is a visiting scholar at Institute of Automation, Chinese Academy of Sciences, and supported by the Brain Research Center, Beijing Institute of Basic Medical Sciences. His research interests include biologically inspired computing and computer vision.
\end{IEEEbiography}
\vspace{-17mm}

\begin{IEEEbiography}[{\includegraphics[width=1in,height=1.2in,clip,keepaspectratio]{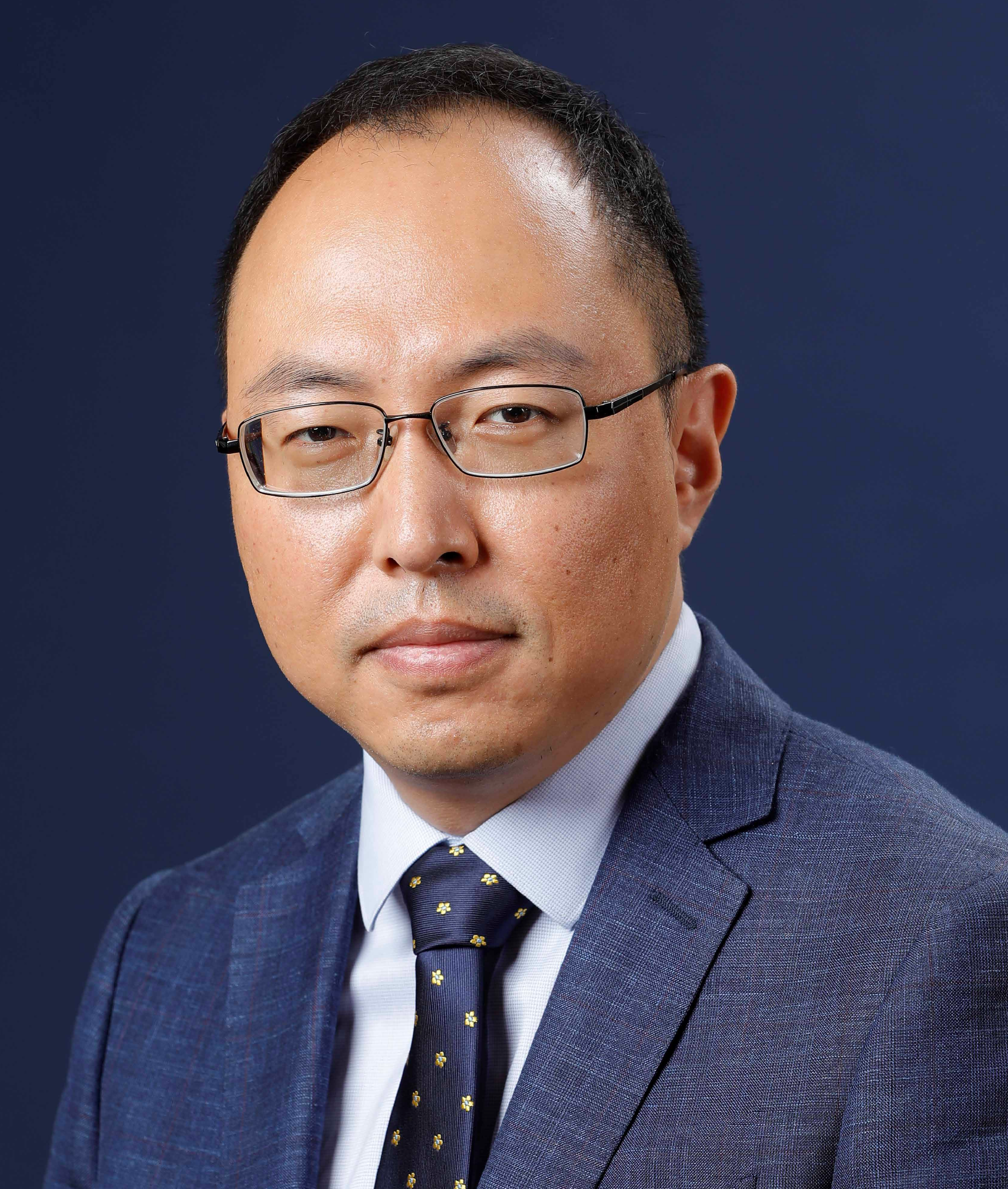}}]{Shan Yu}
	received the BS and PhD degrees in biology from the University of Science and Technology of China, Hefei, China, in 2000 and 2005, respectively. From 2005 to 2014, he conducted Postdoctoral Research with the Max-Planck Institute of Brain Research, Germany (2005–2008), and the National Institute of Mental Health, USA (2008–2014). In 2014, he was with the Institute of Automation, Chinese Academy of Sciences (CASIA), as a recipient of the``One Hundred Talents" program of CAS. He is a Professor with the Brainnetome Center and National Laboratory of Pattern Recognition (NLPR), CASIA. Since 2018, he has been the Deputy Director with the NLPR. He has authored and coauthored more than 30 peer-reviewed papers in neuroscience and other interdisciplinary ﬁelds at lead-ing international journals such as the \textit{Nature Machine Intelligence}, the \textit{Journal of Neuroscience, eLife}, etc. His current research interests include neuronal information processing, brain-inspired computing, and artiﬁcial intelligence.
\end{IEEEbiography}
\vspace{-17mm}

\begin{IEEEbiography}[{\includegraphics[width=1in,height=1.2in,clip,keepaspectratio]{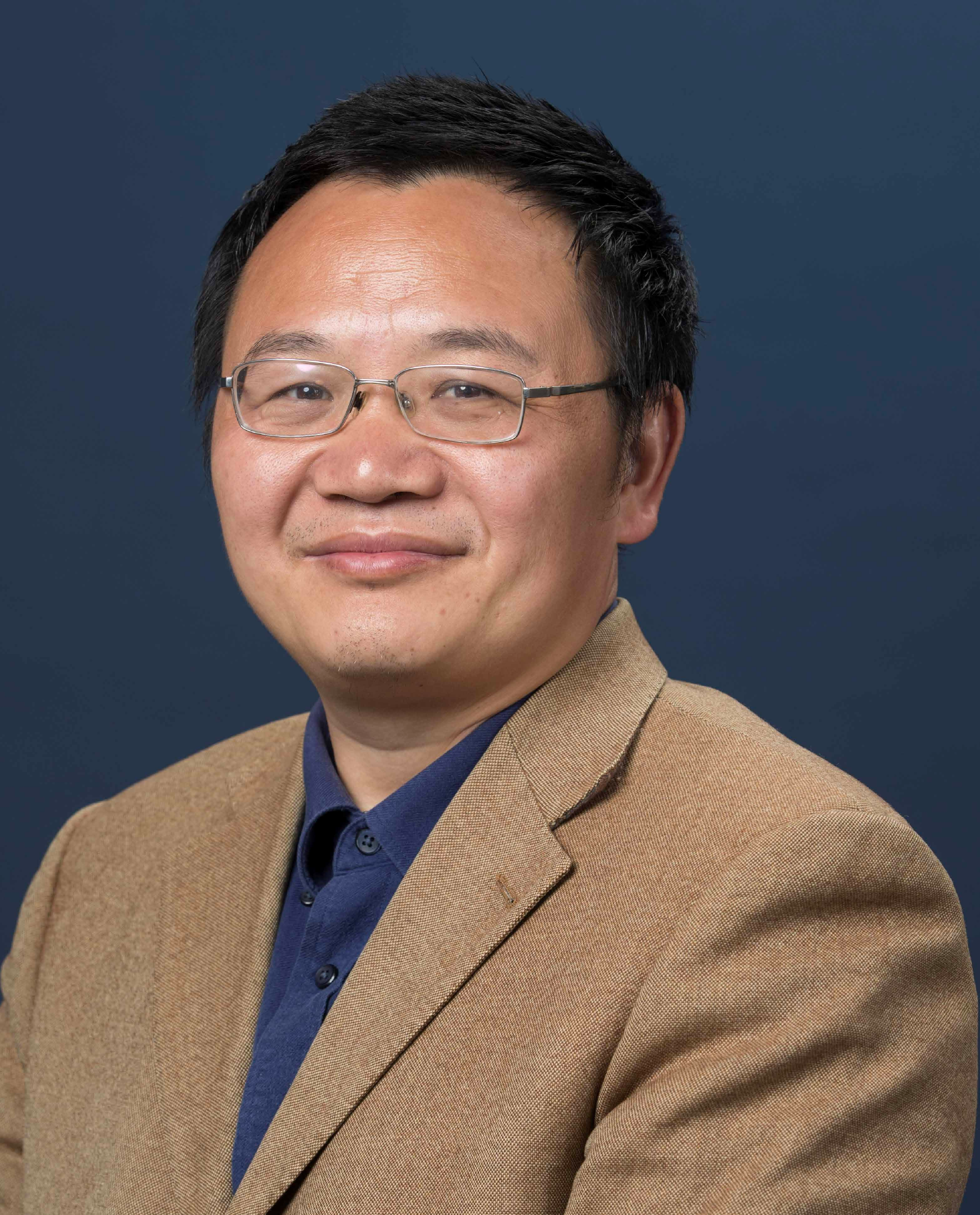}}]{Weiming Hu}
	received the PhD degree from the Department of Computer Science and Engineering, Zhejiang University, Zhejiang, China. Since 1998, he has been with the Institute of Automation, Chinese Academy of Sciences (CASIA), Beijing, where he is currently a Professor. He has published more than 200 papers on peer reviewed international conferences and journals. His current research interests include visual motion analysis and recognition of harmful Internet multimedia.
\end{IEEEbiography}




\clearpage

\begin{center}
\Large\textbf{Supplementary Material}    
\end{center}

This is a supplementary material for our paper ``Recursive Least-Squares Estimator-Aided Online Learning for Visual Tracking". Due to spatial constraints in the main paper, some more detailed explanations with respect to the extension of the derivation of normal equations for MLP in~\cite{Gao2020RLS-RTMDNet} to handle non-linear activation functions, the derivation of normal equations for MLP in the case of cross-entropy loss in RT-MDNet~\cite{Jung2018RTMDNet}, and the use of virtual input in Eqs.~\eqref{equ:mean1}~\eqref{equ:mean2}~\eqref{equ:mean3} of the main paper are moved to this supporting document.

\appendices
\setcounter{equation}{0}
\setcounter{figure}{0}
\renewcommand \thefigure {S\arabic{figure}}
\renewcommand \theequation {S\arabic{equation}}
%
%
%
%

\section{Extending the derivation of normal equations for MLP to handle non-linear activation functions}
In our preliminary version~\cite{Gao2020RLS-RTMDNet}, we have claimed and demonstrated that each layer of MLP in the case of squared-error loss function can be represented by the system of normal equations for RLS estimation by simply ignoring non-linear activation functions. There is no problem for applying the recursive LSE only for the final linear layer. However, without non-linear activation functions, multi-layer MLP becomes just a linear function and there is no reason to have multiple layers. In the real implementation of our RLS-RTMDNet, we actually inherit the non-linear activation functions in the original RT-MDNet as shown in Figure~\ref{fig:MLP}. We hereby provide more explanations on how to extend the derivation of normal equations for MLP in~\cite{Gao2020RLS-RTMDNet} to generally handle other layers including non-linear activation functions. It is noted that we can connect recursive LSE to each MLP layer with non-linear activation and derive the corresponding normal equation individually for different input-output pair $(\mathbf{x}_i, \mathbf{y}_i)$ following the derivation in~\cite{Gao2020RLS-RTMDNet}.

More specifically, we can denote the \textit{squared-error loss} for a $L$-layered MLP neural network with non-linear activation functions using the input-output pair $(\mathbf{x}_i, \mathbf{y}_i)$ as:
\begin{align}
\vspace{-2mm}
\label{equ:lossSI} \mathscr{L}\left(\mathbf{y}_i, \mathbf{W}^{L-1}f^{L-2}(\mathbf{W}^{L-2} \cdots f^2(\mathbf{W}^{2}f^1(\mathbf{W}^{1}\mathbf{x}_i)))\right)~,
\vspace{-2mm}
\end{align}
where the activation functions $f^l(\cdot)$, such as the widely applied ReLU (Rectified Linear Unit) and Leaky ReLU, are \textit{applied to each node separately and the derivative of each activation function is just the diagonal matrix of the derivative on each node}. Backpropagation then gives the gradient of the input values at level $l$ as follows based on the chain rule:
\begin{align}
\vspace{-2mm}
\label{equ:delta1SI} \bm{\delta}^l_i \coloneqq &\left(\mathbf{W}^l\right)^{\top} \cdot diag\left((f^{l})’_i\right) \cdots \left(\mathbf{W}^{L-2}\right)^{\top} \cdot diag\left((f^{L-2})’_i\right)\cdot \nonumber\\ 
&\left(\mathbf{W}^{L-1}\right)^{\top} \cdot \nabla_{\mathbf{z}_i}\mathscr{L}~,
\vspace{-2mm}
\end{align}
where
\begin{align}
\vspace{-2mm}
\label{equ:z1SI} \nabla_{\mathbf{z}_i}\mathscr{L} &= \mathbf{z}_i - \mathbf{y}_i~, \\ 
\label{equ:z2SI}\mathbf{z}_i &= \mathbf{W}^{L-1}f^{L-2}(\mathbf{W}^{L-2} \cdots f^2(\mathbf{W}^{2}f^1(\mathbf{W}^{1}\mathbf{x}_i)))~.
\vspace{-2mm}
\end{align}
The $\bm{\delta}^l_i$ can also easily be computed recursively as:
\begin{align}
\vspace{-2mm}
\label{equ:delta2SI} \bm{\delta}^l_i = \left(\mathbf{W}^l\right)^{\top} \cdot diag\left((f^{l})’_i\right) \cdot \bm{\delta}^{l+1}_i~.
\vspace{-2mm}
\end{align}
The gradient of the weights in layer $l$ is then:
\begin{align}
\vspace{-2mm}
\label{equ:gradient1SI} \nabla_{\mathbf{W}^l}\mathscr{L} = diag\left((f^{l})’_i\right) \cdot \bm{\delta}^{l+1}_i\left(\mathbf{u}^l_i\right)^{\top}~.
\vspace{-2mm}
\end{align}
\textit{Note that the derivative of the weight decay regularization term can be directly added to the result of backpropagation when calculating the gradient of the weights.}

Since the ReLU and Leaky ReLU activation functions are widely applied, we can also use the diagonal matrix of the derivative on each node for them to calculate $\mathbf{z}_i$, \ie,
\begin{align}
\vspace{-2mm}
\label{equ:z3SI}\mathbf{z}_i = &\mathbf{W}^{L-1} \cdot diag\left((f^{L-2})’_i\right) \cdot \mathbf{W}^{L-2} \cdots diag\left((f^{2})’_i\right) \cdot \nonumber\\
&\mathbf{W}^{2} \cdot diag\left((f^{1})’_i\right) \cdot \mathbf{W}^{1}\mathbf{x}_i \\
= &\mathbf{W}^{L-1} \cdot diag\left((f^{L-2})’_i\right) \cdot \mathbf{W}^{L-2} \cdots diag\left((f^{l+1})’_i\right) \cdot \nonumber\\
&\mathbf{W}^{l+1} \cdot diag\left((f^{l})’_i\right) \cdot \mathbf{W}^{l}\mathbf{u}^l_i~.
\vspace{-2mm}
\end{align}
Isolating the term $\mathbf{W}^{l}$ from the rest of the product of the network weight matrices, \ie, let
\begin{align}
\vspace{-2mm}
\label{equ:replaceSI} \mathbf{W}^r_i = &\mathbf{W}^{L-1} \cdot diag\left((f^{L-2})’_i\right) \cdot \mathbf{W}^{L-2} \cdots diag\left((f^{l+1})’_i\right) \cdot \nonumber\\
&\mathbf{W}^{l+1} \cdot diag\left((f^{l})’_i\right)~,
\vspace{-2mm}
\end{align}
we may rewrite
\begin{align}
\vspace{-2mm}
\label{equ:gradient2SI} \nabla_{\mathbf{W}^l}\mathscr{L} &= \left(\mathbf{W}^r_i\right)^{\top}(\mathbf{z}_i - \mathbf{y}_i)\left(\mathbf{u}^l_i\right)^{\top} \\
&= \left(\mathbf{W}^r_i\right)^{\top}(\mathbf{W}^r_i\mathbf{W}^{l}\mathbf{u}^l_i - \mathbf{y}_i)\left(\mathbf{u}^l_i\right)^{\top}~.
\vspace{-2mm}
\end{align}
Solving for $\mathbf{W}^l$ for which $\nabla_{\mathbf{W}^l}\mathscr{L}$ is zero, we may write
\begin{align}
\vspace{-2mm}
\label{equ:gradient3SI} \mathbf{W}^{l}\mathbf{u}^l_i\left(\mathbf{u}^l_i\right)^{\top} = \left(\left(\mathbf{W}^r_i\right)^{\top}\mathbf{W}^r_i\right)^{-1}\left(\mathbf{W}^r_i\right)^{\top}\mathbf{y}_i\left(\mathbf{u}^l_i\right)^{\top} ~.
\vspace{-2mm}
\end{align}
The right side of this new equation can be written as a more intuitive term using some substitution operations, \ie,
\begin{align}
\vspace{-2mm}
\label{equ:substitution} & \left(\left(\mathbf{W}^r_i\right)^{\top}\mathbf{W}^r_i\right)^{-1}\left(\mathbf{W}^r_i\right)^{\top}\left(\mathbf{z}_i  - \nabla_{\mathbf{z}_i}\mathscr{L} \right)\left(\mathbf{u}^l_i\right)^{\top} \nonumber \\ = & \left(\left(\mathbf{W}^r_i\right)^{\!\!\top}\!\mathbf{W}^r_i\right)^{-1}\!\!\left(\left(\mathbf{W}^r_i\right)^{\!\!\top}\!\mathbf{W}^{r}_i\mathbf{z}^{l+1}_i \!-\!diag\!\left((f^{l})’_i\right)\bm{\delta}^{l+1}_i\!\right)\!\!\left(\mathbf{u}^l_i\right)^{\!\!\top} \nonumber \\ = & \left(\mathbf{z}^{l+1}_i - \left(\left(\mathbf{W}^r_i\right)^{\top}\mathbf{W}^r_i\right)^{-1}diag\left((f^{l})’_i\right)\bm{\delta}^{l+1}_i\right)\left(\mathbf{u}^l_i\right)^{\top}~\!\!\!.
\vspace{-2mm}
\end{align}
As it can be seen, this new term includes the desired output $\mathbf{y}^{l+1}_i$ for layer $l+1$, \ie,
\begin{align}
\vspace{-2mm}
\label{equ:substitution} \mathbf{y}^{l+1}_i = \mathbf{z}^{l+1}_i - \left(\left(\mathbf{W}^r_i\right)^{\top}\mathbf{W}^r_i\right)^{-1}diag\left((f^{l})’_i\right)\bm{\delta}^{l+1}_i~,
\vspace{-2mm}
\end{align}
where $\mathbf{z}^{l+1}_i$ is the actual state of the $(l+1)$-th layer's output, and the last term is the ``normalized'' error term. Finally, the system of normal equations for layer $l+1$ is written as:
\begin{align}
\vspace{-2mm}
\label{equ:normalequ} \mathbf{W}^l = \mathbf{y}^{l+1}_i\left(\mathbf{u}^l_i\right)^{\top}\left(\mathbf{u}^l_i\left(\mathbf{u}_i^l\right)^{\top}\right)^{-1}~.
\vspace{-2mm}
\end{align}
The above derivation concludes that each layer of MLP with non-linear activation function can also be represented by the system of normal equations with respect to a specific input-output pair $(\mathbf{x}_i, \mathbf{y}_i)$ for solving the linear LSE problem.

\section{Derivation of normal equations for MLP in RT-MDNet in the case of cross-entropy loss}
\begin{figure*}
\centering
\includegraphics[width=0.90\linewidth]{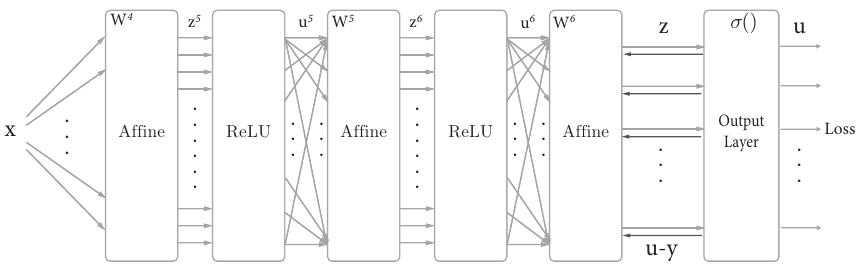}
\caption{The implementation details of the online learned fully-connected MLP layers in RT-MDNet. It is the derivative of the cross-entropy loss $\mathscr{L}_{ce}$ with respect to $\mathbf{z}$, \ie, $\nabla_{\mathbf{z}}\mathscr{L}_{ce} = \mathbf{u} - \mathbf{y}$, that is propagated backwards from higher layers to lower layers. We refer the readers to~\cite{Saitoh2021DLB} for more details.} \label{fig:MLP}
\end{figure*}

\begin{figure*}
\centering
\subfloat[Softmax function]{\includegraphics[width=0.18\linewidth]{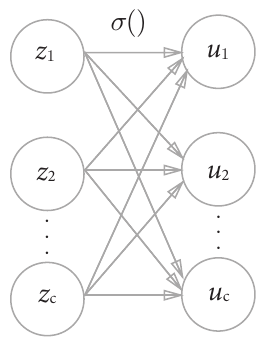}
\label{softmax}}
\hfil
\subfloat[Identity function]{\includegraphics[width=0.18\linewidth]{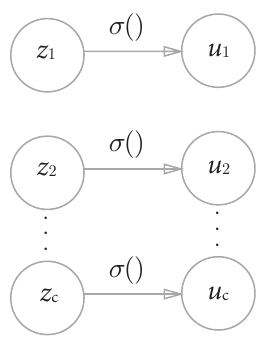}
\label{identity}}
\caption{Different designs for the output layer: Softmax function for the classification problems and Identity function for the regression problems.} \label{fig:outputlayer}
\end{figure*}
According to~\cite{Saitoh2021DLB}, the product of matrices in forward propagation in a MLP neural network is called an ``affine transformation" in the field of geometry. Therefore, the process that performs an affine transformation can be implemented as an ``Affine layer" in  RT-MDNet~\cite{Jung2018RTMDNet}. In Figure~\ref{fig:MLP}, we show the implementation details of the online learned fully-connected MLP layers (\verb'fc4-6') $\left\{\mathbf{W}^l\right\}_{l=4}^6$. In RT-MDNet, the output scores $\mathbf{z}$ of the single fully-connected binary classification layer $\mathbf{W}^6$ are firstly normalized by a Softmax output layer and then converted to the probability values $\mathbf{u}$ of belonging to different classes. Finally, the cross-entropy error function is applied for calculating loss for online training. It can also be derived that it is the derivative of the cross-entropy loss $\mathscr{L}_{ce}$ with respect to $\mathbf{z}$, \ie, $\nabla_{\mathbf{z}}\mathscr{L}_{ce} = \mathbf{u} - \mathbf{y}$, that is propagated backwards from higher layers to lower layers. We refer the readers to~\cite{Saitoh2021DLB} for more details.

As~\cite{Saitoh2021DLB} has stated, using the cross-entropy loss function for the Softmax output layer (Figure~\ref{softmax}) makes backward propagation return the ``pretty" result $\nabla_{\mathbf{z}}\mathscr{L}_{ce} = \mathbf{u} - \mathbf{y}$, which is not accidental; the cross-entropy error function is designed to do this, especially for the classification problems. In the regression problems, the sum of squared errors is used for calculating loss for the same reason as the output layer can be seen as an Identity function (Figure~\ref{identity}) and backward propagation returns the same ``pretty" result $\nabla_{\mathbf{z}}\mathscr{L}_{se} = \mathbf{u} - \mathbf{y}$ for the squared-error loss $\mathscr{L}_{se}$. That means all the affine layers in the different cases of cross-entropy loss and squared-error loss will receive the gradient information from the same backpropagation of $\mathbf{u} - \mathbf{y}$ for the same input-output pair $(\mathbf{x}, \mathbf{y})$.  This is the basic reason why we claimed in our preliminary conference version~\cite{Gao2020RLS-RTMDNet} that all the normal equation derivations for RLS estimation in the case of squared-error loss function can be easily transplanted to the case of cross-entropy loss.

More specifically, although we can not directly use the cross-entropy loss for deriving the normal equations in RT-MDNet due to the special form of Softmax-based output layer\footnote{It is very different from the ReLU-based non-linear activation layer where the activation function is applied to each node separately.}, we can instead cast the cross-entropy loss along with its Softmax-based output layer as a black box component, and let the affine layers believe they are receiving the gradient information from a squared-error loss with the Identity function as the output layer. In other words, the update of the affine layers will behave like in the case of squared-error loss to incorporate memory retention.

\section{Explanation of using virtual input}

We provide more detailed explanations about the virtual input defined in Eqs.~\eqref{equ:mean1}~\eqref{equ:mean2}~\eqref{equ:mean3} of the main paper as follows. We start by providing the proof of Eq.~\eqref{equ:submultiplicativity}, which means the minimization of Eq.~\eqref{equ:costfunctionLSnew2} will also lead to the minimization of the following {\it cost function}:
\begin{align}
\label{appequ:costfunctionLSnew5} \mathscr{L}(\mathbf{W}_n) = \mathop{\sum}\limits _{i=1}^{n}\beta^{n-i}\!\left\Vert \overline{\mathbf{y}}_i - \mathbf{W}_n\overline{\mathbf{x}}_i\right\Vert^2 + \frac{\delta}{b}\beta^n\left\Vert \mathbf{W}_n\right\Vert^2~.
\end{align}
This helps to explain why it is feasible to update $\mathbf{P}_{n}$ approximately based on the virtual input in the case of $b > 1$.

\textit{Proof of Eq.~\eqref{equ:submultiplicativity}.} Based on the Cauchy-Schwarz inequality, it holds that
\begin{align}
\label{appequ:submultiplicativity} \left\Vert \mathop{\sum}\limits _{j=1}^{b}\left(\mathbf{y}_{i,j} - \mathbf{W}_n\mathbf{x}_{i,j}\right)\right\Vert^2 \leq b\mathop{\sum}\limits _{j=1}^{b}\left\Vert \mathbf{y}_{i,j} - \mathbf{W}_n\mathbf{x}_{i,j}\right\Vert^2~,
\end{align}
for all the $b$ samples in the $i_{th}$ {\it block}. Using Eq.~\eqref{equ:mean1}, we can derive the following inequality
\begin{align}
\label{appequ:inequality} \left\Vert \mathop{\sum}\limits _{j=1}^{b}\left(\mathbf{y}_{i,j} - \mathbf{W}_n\mathbf{x}_{i,j}\right)\right\Vert^2 &= b^2\left\Vert \overline{\mathbf{y}}_i - \mathbf{W}_n\overline{\mathbf{x}}_i\right\Vert^2\nonumber\\
&\leq b\mathop{\sum}\limits _{j=1}^{b}\left\Vert \mathbf{y}_{i,j} - \mathbf{W}_n\mathbf{x}_{i,j}\right\Vert^2
\end{align}
to have
\begin{align}
\label{appequ:inequality2} \left\Vert \overline{\mathbf{y}}_i - \mathbf{W}_n\overline{\mathbf{x}}_i\right\Vert^2 \leq \frac{1}{b}\left\Vert \bm{Y}_i - \bm{X}_i\mathbf{W}_n^{\top}\right\Vert^2~.
\end{align}

\textit{Explanation of Eq.~\eqref{equ:mean2} in Section~\ref{Sec:MLP}.} In the real implementation of RT-MDNet, it is the average of all derivatives of the loss function for all the $b$ samples in the mini-batch that is used for updating $\mathbf{W}^l$ at each optimization iteration of the MBSGD process. Since the recursive formula for improving this process has a similar expression of the {\it cost function} to Eq.~\eqref{equ:costfunctionLSnew2}, we thus can directly define the virtual input for each MLP layer in RT-MDNet at the time index $n$ according to Eq.~\eqref{equ:mean1} as follows:
\begin{align}
\label{appequ:mean2} \overline{\mathbf{x}}_n^l = \frac{1}{b}\mathop{\sum}\limits _{j=1}^{b}\mathbf{x}_{j}^l~.
\end{align}

\textit{Explanation of Eq.~\eqref{equ:mean3} in Section~\ref{Sec:CNN}.} In the real implementation of DiMP, it is the sum of all derivatives of the squared-error loss in Eq.~\eqref{equ:costfunctionLSATOM3} for all the $N$ samples that is used for updating $\mathbf{W}$ at each optimization iteration of the BGD process. According to a variation of Eq.~\eqref{equ:submultiplicativity}, \ie,
\begin{align}
\label{appequ:inequality3} \left\Vert \sqrt{b}\overline{\mathbf{y}}_i - \mathbf{W}_n\left(\sqrt{b}\overline{\mathbf{x}}_i\right)\right\Vert^2 \leq \mathop{\sum}\limits _{j=1}^{b}\left\Vert \mathbf{y}_{i,j} - \mathbf{W}_n\mathbf{x}_{i,j}\right\Vert^2~,
\end{align}
we can define the virtual input for the convolutional layer in DiMP at the time index $n$ as follows:
\begin{align}
\label{appequ:mean3} \overline{\mathbf{x}}_n = \frac{\sqrt{NM}}{NM} \mathop{\sum}\limits _{j=1}^{N}\mathop{\sum}\limits _{k=1}^{M}\sqrt{\gamma_{jk}}\mathbf{x}_{jk} = \frac{1}{\sqrt{NM}} \mathop{\sum}\limits _{j=1}^{N}\mathop{\sum}\limits _{k=1}^{M}\sqrt{\gamma_{jk}}\mathbf{x}_{jk}.
\end{align}

\end{document}